\documentclass{article}

\usepackage{arxiv}

\usepackage[utf8]{inputenc} 
\usepackage[T1]{fontenc}    
\usepackage{hyperref}       
\usepackage{url}            
\usepackage{booktabs}       
\usepackage{amsmath,amssymb,amsfonts}       
\usepackage{nicefrac}       
\usepackage{microtype}      
\usepackage{lipsum}
\usepackage{graphicx}
\usepackage{algorithmic}
\usepackage{algorithm}

\title{Deep Stacked Stochastic Configuration Networks for Lifelong Learning of Non-Stationary Data Streams \thanks{This paper has been published in Information Sciences.}}

\author{
  Mahardhika Pratama\\
  School of Computer Science and Engineering\\
  Nanyang Technological University\\
  Singapore\\
  \texttt{mpratama@ntu.edu.sg} \\
  \And
  Dianhui Wang\\
  The State Key Laboratory of Synthetical Automation for Process Industries\\
  Northeastern University\\
  China\\
  \texttt{dh.wang@latrobe.edu.au}
}

\begin{document}
\maketitle

\begin{abstract}
The concept of SCN offers a fast framework with universal approximation guarantee for lifelong learning of non-stationary data streams. Its adaptive scope selection property enables for proper random generation of hidden unit parameters advancing conventional randomized approaches constrained with a fixed scope of random parameters. This paper proposes deep stacked stochastic configuration network (DSSCN) for continual learning of non-stationary data streams which contributes two major aspects: 1) DSSCN features a self-constructing methodology of deep stacked network structure where hidden unit and hidden layer are extracted automatically from continuously generated data streams; 2) the concept of SCN is developed to randomly assign inverse covariance matrix of multivariate Gaussian function in the hidden node addition step bypassing its computationally prohibitive tuning phase. Numerical evaluation and comparison with prominent data stream algorithms under two procedures: periodic hold-out and prequential test-then-train processes demonstrate the advantage of proposed methodology.
\end{abstract}


\section{Introduction}
Although issue of randomized approaches in neural networks have received attention in the late 80's and early 90's, the benefit of such approaches becomes more evident in recent days with the emergence of deep neural networks (DNNs) involving a large number of network parameters where training all of those parameters incurs considerable computational cost and prohibitive memory demand. The randomized approach relieves design steps of DNNs because it allows most DNN parameters except the output weights to be randomly generated from certain scopes and distributions. This mechanism speeds up algorithm's runtime which happens to be a key factor in the resource-constrained environments such as data streams because the training process rely on no tuning of hidden nodes and is free from iterative training process. Random vector functional link network (RVFLNs) exemplifies a successful integration of the randomized approaches in neural network \cite{pao1992functional,pao1994learning}. Specifically, RVFLN incorporates the randomized idea for construction of functional link neural network \cite{pao1992functional} and shares the universal approximation capability \cite{igelnik1995stochastic}. It is, however, reported that the universal approximation capability of RVFLN highly depends on the setting of the scope of random parameters and is hardly implemented in practise \cite{igelnik1995stochastic}. It is shown in \cite{li2017insights} that the scope of random parameters plays vital role to retain the universal approximation property of RVFLN. The scope of random parameters should not be fixed and should be adaptively adjusted to adapt to different problem complexities \cite{SCN}. 

Stochastic configuration networks (SCN) were recently proposed in \cite{SCN} as an innovative and revolutionary solution of randomized learning with guaranteed universal approximation capability. It combats the bottleneck of conventional randomized learning where random parameters are generated from blindly selected and static scopes. SCN applies a supervisory mechanism to assign the random parameters rendering the scope of random parameters data-dependent and flexible. In \cite{SCN}, three algorithms to implement SCNs are proposed by tuning the output weights {red}{in three configurations}: direct, local least square and global least square. The SCN framework has been extended in \cite{DeepSCN} where the SCN theory is implemented in the context of deep neural networks (DNNs). Ensemble version of SCN was proposed in \cite{SCNensemble} which can be regarded as an advancement of Decorrelated Neural Network Ensemble (DNNE) \cite{alhamdoosh2014fast}. It resolves the problem of large-scale pseudo-inverse computation using the block Jacobi and Gauss-Seidel approximation. A robust SCN was designed in \cite{robustSCN} for resolving robust data modeling problems using the kernel density estimation. {While SCN has potency for solving data stream problems since it is capable of significantly relieving parameter tuning efforts}, data stream remains an open issue for these existing SCNs because they work on non-continual setting and learning environment is supposed to be unchanged during the course of training process.

In this paper, we propose a novel version of deepSCNs termed deep stacked stochastic configuration networks (DSSCNs) which aims to solve two key issues: 1) \textbf{Simplification of Data Stream Algorithm}. DSSCN embraces the concept of SCN to set the hidden node parameters during addition of new hidden node procedure without further tuning process. The SCN concept enables random generation of hidden node parameters with analytically determined scopes while retaining universal approximation property. Note that {existing evolving intelligent systems (EISs) \cite{EFSSURVEY} normally utilize computationally expensive hidden node tuning process: extended kalman filter (EKF), stochastic gradient descent (SGD) \cite{juang1999recurrent}, or clustering approach \cite{lughofer2008flexfis,GENSMARTEFS}, etc.} {This approach} generalizes evolving random vector functional link network (eRVFLN) \cite{et2rvfln} still operating under the scope of randome parameters $[-1,1]$; 2) {\textbf{Automatic Construction of Deep Stacked Network Structure from Data Streams}.} 
DSSCN demonstrates an open structure principle where not only hidden nodes are automatically evolved and pruned from data streams but also hidden layers of DSSCN are self-organized. In other words, the selection of DSSCN's depth is fully data-driven {in which its hidden layer is capable of growing and shrinking} \cite{pENsemble}. Our algorithm is capable of initiating its learning process from scratch with no initial structure. That is, hidden node and hidden layer can be incrementally added from data streams subject to the rule growing scenario and the drift detection method.

DSSCN is built upon the stacked generalization principle \cite{stackedgeneralization}. That is, the output of one learner is fed as an input to the next learner while each layer is given with the same input representation plus a random projection of the predictive output of the preceding layer as developed in \cite{stackedDFNN}. A drift detection scenario controls the depth of the DSSCN with addition of a new layer if a drift is identified. DSSCN implements complexity reduction mechanisms in terms of online feature weighting mechanism and hidden layer pruning procedure. Salient characteristic of DSSCN is outlined as follows:
\begin{enumerate}
\item \textit{Drift Detection Scenario}: the evolving structure of DSSCN is governed by a drift detection scenario in which a new layer is incorporated by constructing a new hidden layer if a drift is signalled. Drift detection scenario has been widely implemented in realm of ensemble learning where it controls when a new ensemble member is added \cite{pENsemble}. Nevertheless, recent investigation in \cite{powerofdepth} has concluded that the increase of network depth make significant difference in generalization power. It is theoretically shown that a two layer neural network cannot solve a task that can be addressed by a three layer neural network unless it has an infinite number of hidden nodes. Our work aims to investigate the use of the drift detection scenario to signal {the increase of network depth}. The drift detection scenario is devised from the Hoeffding's bound theory which categorizes data streams into three conditions: drift, warning and {normal \cite{drift}.} The idea of Hoeffding's bound allows the choice of conflict threshold to be determined automatically with sufficient statistical guarantee. The underlying rationale behind the deep approach for {processing} data streams is outlined in {Section 7} and numerically benchmarked against a fixed-structured DNN.
\item \textit{Evolving Stochastic Configuration Networks}: evolving stochastic configuration networks (eSCNs) is crafted in this paper where it is derived from evolving random vector functional link networks (eRVFLNs)\cite{et2rvfln}. Unlike its predecessor, eSCNs adopt the SCNs theory where inequality constraint is deployed when initializing parameters of a new rule \cite{SCN}. The concept of SCNs allows adaptive scope allocation where the scope of random parameters are not fixed to a specific range rather are dynamically selected. As with eT2RVFLN, the network output is inferred using the interval type-2 fuzzy system concept where interval type-2 multivariate Gaussian function with uncertain centroids is integrated. The use of multivariate Gaussian rule reduces fuzzy rule demand due to its non-axis-parallel property but it incurs high computational and storage complexity due to the presence of non-diagonal covariance matrix. The SCN concept addresses this problem because it is capable of skipping the hidden unit tuning step while still guaranteeing the universal approximation property \cite{SCN}. Moreover, the concept of eSCN is applicable to the type-1 fuzzy system as well with some minor modifications.
\item \textit{Deep Stacked Neural Networks Architecture}: DSSCN is actualized under a deep stacked neural networks structure where every layer is formed by a base-learner, eSCNs. The stacked neural network architecture is built upon the stacked generalization principle where the training samples of every layer are drawn from original training samples mixed with random shifts \cite{stackedDFNN}. Random shift is induced by random projection of predictive outcomes of preceding layer/learner. This idea is inspired by the deep TSK fuzzy learners in \cite{stackedDFNN}. Here, we go one step further by introducing the self-organizing concept of DNN's structure encompassing hidden nodes, input features and depth of the networks. 
\item \textit{Online Layer Pruning Scenario}: DSSCN is equipped with a complexity reduction method using the learner pruning strategy. The learner pruning strategy analyzes the output of base learners and two base learners sharing high similarity are coalesced to alleviate complexity and risk of overfitting. The learner pruning scenario is enabled in the DSSCN without the risk of catastrophic forgetting because it features a stacked architecture where the output of base learner is utilized as a shifting factor to the next base learner. In other words, each layer is supposed to output the same predicted variables. 
\item \textit{Online Feature Weighting Scenario}: DSSCN features the online feature weighting mechanism based on the notion of feature redundancy \cite{pratama2016scaffolding}. This mechanism deploys feature weights where features exhibiting high mutual information is assigned with low weights. Moreover, feature weights are not fixed and dynamically tuned as the redundancy level of input attributes.
\end{enumerate}

Learning performance of DSSCN has been evaluated by exploiting five popular {streaming problems} and one real-world application from our own project, RFID based Indoor localization. It is also compared with six data stream algorithms, pClass \cite{pratama2015pclass}, eT2Class \cite{pratama2016evolving}, pEnsemble \cite{pENsemble}, Learn++CDS \cite{DitzlerImbalanced}, Learn++NSE \cite{Leanr++NSE} and its shallow version, eSCN. DSSCN demonstrates improvement of classification rates over its counterparts while imposing comparable computational power and network parameters. Compared to the shallow version, 2\% to 5\% improvement of predictive accuracy can be achieved with {extra computational} and storage complexity. Moreover, DSSCN has been numerically validated under the prequential test-then-train protocol and compared with a static DNN where it demonstrates more encouraging performance in terms of predictive accuracy than standard DNN.

The remainder of this paper is organized as follows: Section 2 reviews related works of deep neural networks; Section 3 elaborates the problem formulation of data stream mining under two protocols, prequential test-then-train and periodic hold-out processes; Section 4 discusses the network architecture of DSSCN; Section 5 describes the learning scheme of DSSCN: online feature weighting mechanism, hidden layer pruning mechanism, hidden layer growing mechanism and evolving stochastic configuration networks; Performance evaluation is discussed in Section 6; Section 7 outlines technical reasons behind the use of deep structure in handling data streams; Some concluding remarks are drawn in Section 8.

\section{Related Works in Deep Neural Networks}
Deep neural network (DNN) has gained popularity in the recent days since it delivers state-of-the art performance compared to conventional machine learning variants. It is proven theoretically in \cite{powerofdepth} that the depth of neural network structure highly influences its generalization power and major performance improvement can be attained by adding the depth of DNN over the width of DNN i.e., inserting hidden nodes. Notwithstanding that it outperforms other machine learning methods, DNN's training process draws significantly higher cost than its shallow counterpart. Since DNNs still rely on blind model selection where the network architecture is determined purely from expert domain knowledge or trial-error mechanism, one often arrives at much higher network complexity than what necessary. This situation ensues because a very deep network architecture is well-known to mostly produce decent approximation power. Such DNN is evident to be at a higher risk of complicated training process due to exploding or vanishing gradients and to incur expensive memory requirement and slow computational speed. Because of a high number of network parameters, DNNs necessitate abundant training samples to ensure parameter and error convergence.

Model selection has been a major research issue in DNNs and various contributions have been devoted to propose effective model selection scenarios encompassing pruning method \cite{deeplinear}, regularization \cite{CirculantProjections}, parameter prediction \cite{parameterprediction}, etc. In \cite{LearningTheNumber}, a group sparsity regularization technique has been proposed to relieve the network complexity of DNN where each group is treated as a single neuron. It starts from an over-complete network and complexity reduction is carried out by forcing network parameters to zero. This concept is substantiated by the finding of \cite{parameterprediction} where there exists significant redundancy in the parameterizations of DNNs. It is demonstrated that similar performance of DNN can be generated using only fewer network parameters. DNN with a stochastic depth was proposed in \cite{stochasticdepth} where the salient contribution is the introduction of random subset selection. That is, the subset of hidden layers are randomly dropped and bypassed with an identity function. The parameter redundancy of DNN is overcome using the concept of circular projection \cite{CirculantProjections}. The circular projection method makes use of circulant matrix which reduces the number of network parameters. The concept of deep structure has been adopted in the context of fuzzy neural networks (FNNs). In \cite{stackedDFNN}, a slightly different approach was proposed where the random projection concept is utilized to realize a deep network structure. Our algorithm is inspired by these works but we go one step ahead by proposing an automated and online approach in the construction of network structure - hidden layer, hidden unit, input feature are evolved. Online learning of DNN was also proposed in \cite{OnlineDeepLearning}. It is as with conventional model selection approaches where an over-complete DNN is built initially and complexity reduction scenario through the hedging concept is applied afterward to drop unimportant components of DNN's structure in the classification process. It is noted that the use of DNNs for mining data streams remains an open issue notably due to their offline learning process and lack of adaptation to rapidly changing environments.

\section{Problem Formulation}
Data stream is a special problem of machine learning field where data points are continuously generated in the lifelong fashion ${C_k=[C_1,C_2,...,C_K]}$ where ${C_k}$ can be in the form of data batch ${C_k}\in\Re^{N\times n}$ or a single data point. $K$ denotes the number of time stamp unknown before process runs. In practise, data stream arrives with the absence of true class label where ${C_k=X_k}\in\Re^{N\times n}$ and operator labelling effort is solicited to assign the true class labels $Y_k\in\Re^{N}$. The $0-1$ encoding scheme can be applied to construct the class matrix ${Y_k\in\Re^{N\times m}}$. There exist two commonly adopted simulation procedures of data streams, namely \textbf{Periodic Hold Out} and \textbf{Prequential}. The key difference lies in the order of training and testing processes, train-then-test (periodic hold-out) and test-then-train (prequential). The typical characteristic of data stream is perceived in the presence of \textbf{concept drift} defined as the change of posterior probability $P(Y|X)_{K}\neq P(Y|K)_{K-1}$. Overview of DSSCN's learning procedures are visualized in Figure \ref{Periodic_Hold_Out} and \ref{Prequential_Test_Then_Train} under {the periodic hold-out and prequential test-then-train simulation procedures.}

DSSCN is proposed here to address the data stream problem where it realizes a deep elastic structure built upon the stacked generalization principle \cite{stackedgeneralization}. This structure adopts the concept of random shift \cite{stackedDFNN} where every hidden layer or classifier is connected in series and accepts the original input pattern influenced by random shift. DSSCN realizes the concept of dynamic depth where the hidden layer can be dynamically generated in the presence of drift. This strategy is motivated by the fact that addition of hidden layer is more meaningful than addition of hidden units or classifiers \cite{powerofdepth} simply because it enhances the network capacity more significantly than addition of hidden units or classifiers \cite{deeplinear}. The concept of dynamic depth is further actualized using the hidden layer pruning procedure which discards layers sharing high mutual information. To improve its flexibility further, the hidden layer of DSSCN characterizes an open structure principle which is capable of growing and pruning hidden nodes on demand and is combined by the SCN theory to produce dynamic scopes of random parameters \cite{SCN}.  
\begin{center}
	\begin{figure}\label{Periodic_Hold_Out}
		\begin{centering}
			\includegraphics[scale=0.4]{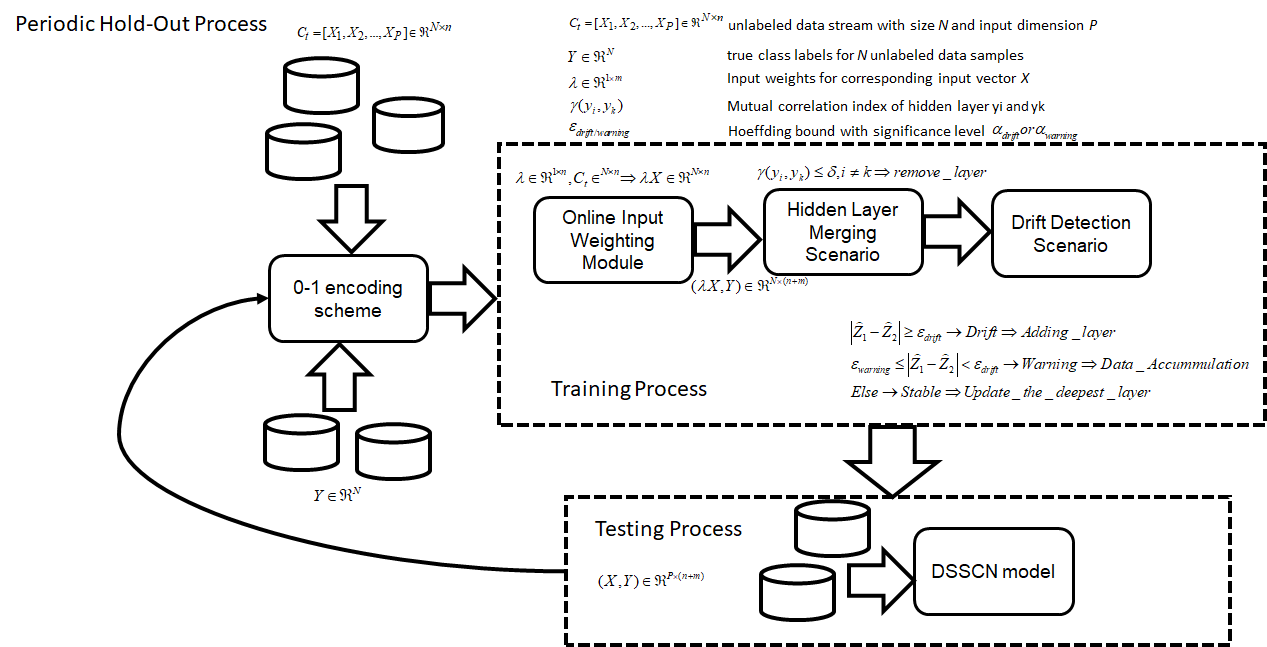}
			\par\end{centering}
		\caption{{Problem Formulation: Periodic Hold Out Process}}
	\end{figure}
\par\end{center}
\begin{center}
	\begin{figure}\label{Prequential_Test_Then_Train}
		\begin{centering}
			\includegraphics[scale=0.4]{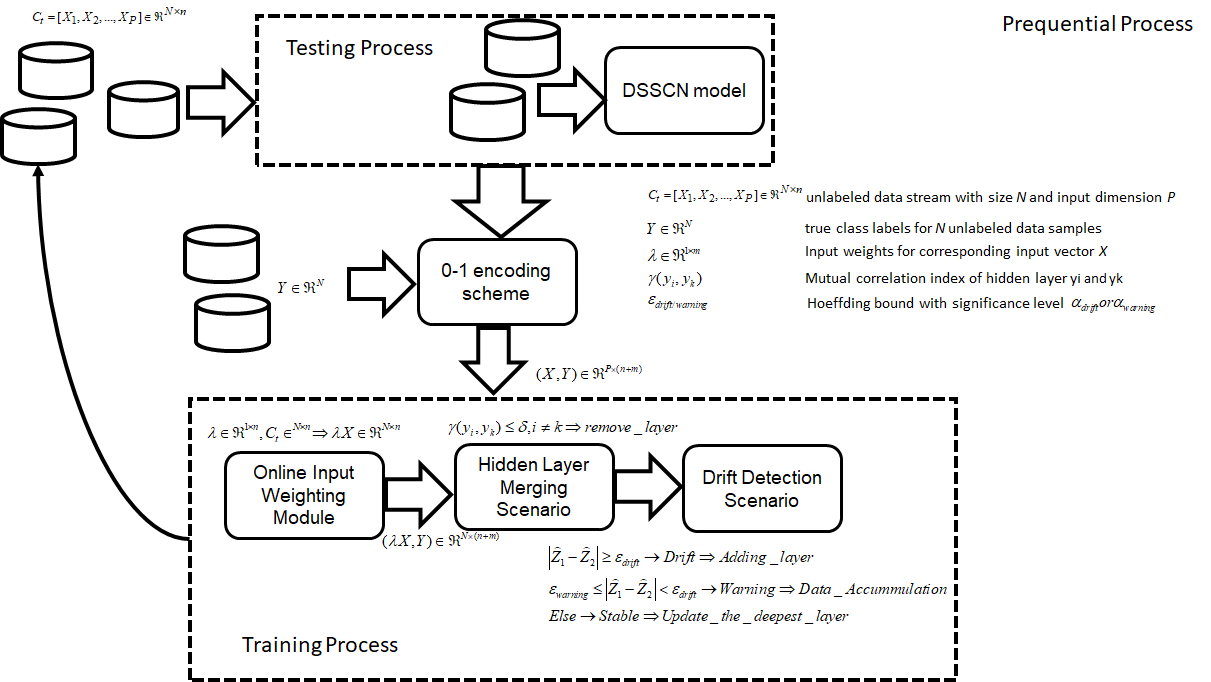}
			\par\end{centering}
		\caption{{Problem Formulation: Prequential Test Then Train Process}}
	\end{figure}
\par\end{center}
\begin{table}[t]
		\caption{List of Symbols}
		\begin{centering}
		\label{tab:Network Parameters}
		\scalebox{0.8}{
		\begin{tabular}{clc}
		\toprule
		\textbf{Symbols} & \textbf{Dimension} & \textbf{Definition}\tabularnewline
		\midrule
    		${C_k}$ & $N\times n$ & unlabelled data batch \tabularnewline
    		$K$ & the number of data batches & N/A \tabularnewline
    		$N$ & the batch size & N/A \tabularnewline
    		$n$ & input dimension & N/A \tabularnewline
    		${X_k}$ & input data batch & $N\times n$ \tabularnewline
    		${Y_k}$ & target data batch & $N\times m$ \tabularnewline
    		$m$ & output dimension & N/A \tabularnewline
    		$P(Y|X)$ & output posterior probability & N/A \tabularnewline
    		$\lambda$ & input weight & $1\times n$ \tabularnewline
    		$\alpha$ & random projection constant & N/A \tabularnewline
    		${P_{D-1}}$ & random projection matrix of $D-1$ layer & $m\times n$ \tabularnewline
    		$D$ & the number of hidden layers & N/A \tabularnewline
    		$\underline{C_D}$ & the lower center of interval-valued activation function of $D-th$ hidden layer & $R\times n$ \tabularnewline
    		${\overline{C_D}}$ & the upper center of interval-valued activation function of $D-th$ hidden layer & $R\times n$ \tabularnewline
    		${\sum^{-1}}$ & the inverse covariance matrix of interval-valued activation function of $D-th$ hidden layer & $R(n\times n)$ \tabularnewline
    		$R$ & the number of hidden nodes & N/A \tabularnewline
    		${W_D}$ & the output weight matrix of $D-th$ layer & $R(2n+1)\times m$ \tabularnewline
    		${\Omega_i}$ & the output covariance matrix of $i-th$ layer & $(2n+1)\times (2n+1) \times m$ \tabularnewline
    		$\gamma(x_1,x_2)$ & mutual compression index of two variables $x_1,x_2$ & N/A \tabularnewline
    		$\epsilon_{\alpha}$ & Hoeffding bound with the significant level $\alpha$ & N/A \tabularnewline
    		$\alpha_D$ & significance level of drift level & N/A \tabularnewline
    		$\alpha_W$ & significance level of warning level & N/A \tabularnewline
    		$cut$ & cutting point of drift detection phase & N/A \tabularnewline
    		$\overline{G_i}$ & upper firing strength of $i-th$ node & N/A \tabularnewline
    		$\underline{G_i}$ & lower firing strength of $i-th$ node & N/A \tabularnewline
    		$q$ & design coefficient & N/A \tabularnewline
    		$[-\xi,\xi]$ & dynamic scope of random parameters & N/A \tabularnewline
		\bottomrule
				
		\end{tabular}}
		\par\end{centering}
	\end{table}

\begin{center}
	\begin{figure}\label{learning architecture of DSSCN}
		\begin{centering}	\includegraphics[scale=0.45]{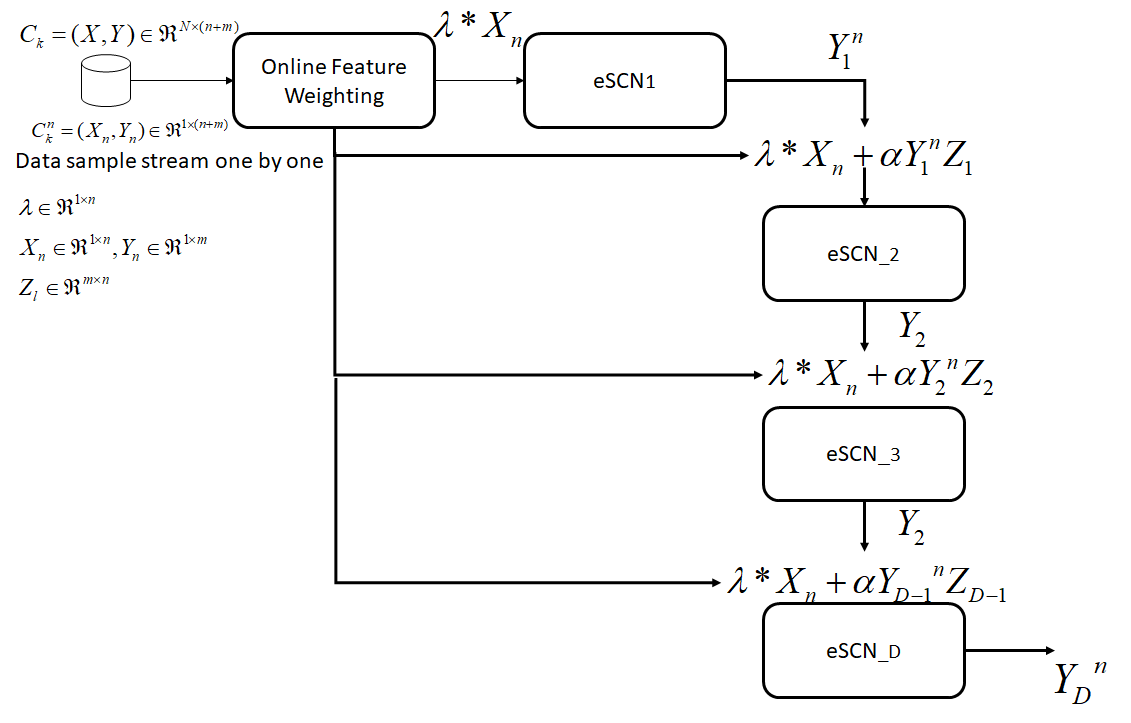}
			\par\end{centering}
		\caption{Network Architecture of DSSCN}
	\end{figure}
\par\end{center}

\section{Architecture of Deep Stacked Stochastic Configuration Networks}
Deep stacked architecture of DSSCNs is inspired by the work of \cite{stackedDFNN} featuring a series of base building units working in tandem.
That is, each base building unit is structured by eSCNs which receives
the original input attributes added by the random shift. The novelty
of DSSCNs lies in its autonomous working principle where the deep stacked
architecture can be automatically deepened and shallowed. The random
shift is induced by the random projection of the predictive output
of the preceding layer which aims to follow the stacked
generalization principle \cite{stackedgeneralization}. The deep stacked architecture allows continuous
refinement of predictive output through multiple nonlinear mapping
of base building units. The architecture of deep stacked structure
is depicted in Figure \ref{learning architecture of DSSCN}. DSSCN works in chunk by chunk basis where
each chunk containing $N$ pairs of tupples $(X_{t},Y_{t})_{t=1,..,N}$ streams
overtime and the total number of chunks is unknown. It is worth mentioning
that DSSCN adopts a strictly single-pass learning process where each
data chunk is discarded once learned. In realm of deep stacked network
architecture, it is challenging to adopt the sample by sample
update since the hierarchical structure of deep stacked network relies
on the output of the preceding layer. That is, the output of the preceding
layer must feed meaningful information for the next layer to lessen
the predictive error toward zero.

The first building block or layer is injected with the original data
chunk $X_{1}=(X_{t})_{t=1,...,N}$ weighted by the input weight vector $\lambda\in\Re^{n}$ and in turn generates the predictive
outputs $Y_{1}=(Y_{t})_{t=1,...,N}$.The stacked generalization principle
is implemented in the next layer where the output of the first building
unit is connected to the second layer and mixed with the random projection
matrix ${\boldsymbol{P}}$. The input of the second building block is formulated as follows:
\begin{equation}
{
X_{2}=\lambda X+\alpha Y_{1}\boldsymbol{P_{1}},
}
\end{equation}
where $\alpha$ is a random projection constant and ${\boldsymbol{P_{1}}\in\mathbb{R}^{m\times n}}$ is
the random projection matrix randomly generated in the range of {[}0,1{]} while $\lambda$ is the input weight vector crafted as Section 6.1
It is observed in {(1)} that the right term $\alpha Y_{1}\boldsymbol{Z_{1}}$is
treated as the random shift of the original input attributes which
aims to capture a high level abstraction of the original input features.
The stacking strategy continues up to the last building block whose
input is determined as ${X_{D}=\lambda X+\alpha Y_{D-1}\boldsymbol{P_{D-1}}}$ where $D$
stands for the depth of DSSCN. It is worth noting that DSSCN demonstrates
the fully elastic deep network structure where a new building unit
can be inserted on top of the current hidden layer when the existing structure
does not suffice to a given problem. All layers or building blocks
are stacked and work in tandem where each layer except the bottom layer produces random shifts of
original input pattern to be passed to the next layer. 

It is evident that the notion of random shifts in each layer of DSSCN
improves linear separability of given classification problem because
it paves a way to move apart the manifolds of original problem in
a stacked manner \cite{stackedDFNN}. This structure is also perceived as a sort
of hierarchical structure but it differs from the conventional hierarchical deep neural network which generally loses transparency of intermediate features because it can no longer be associated with physical semantics of original input features \cite{stackedDFNN3}. It is also worth mentioning that DSSCN is equipped by the {online} feature selection mechanism coupled in every building unit. This mechanism selects relevant input attributes for every layer which enhances generalization power.
\begin{algorithm}
    \caption{{Architecture of DSSCN}}
    \label{dsscn}
    \begin{algorithmic}[1]
        	\STATE \textit{Input}: Data Chunk$(\boldsymbol{X},\boldsymbol{Y})\in\mathbb{R^{\mathrm{N\times(n+m)}}}$, the layer pruning threshold$\delta$, \\the time-constants of the Hoeffding's based drift detection method$T_{D},T_{W}$, the random projection constant $\alpha$
		    \STATE \textit{Output}: Predicted Class labels $\hat{Y}\in\mathbb{R^{\mathrm{N\times(n+m)}}}$,DSSCN structure with $D$ depth or layers\\ $f(layer_{1}(\boldsymbol{\widetilde{C}}=[\underline{C},\overline{C}]\in\mathbb{R^{\mathrm{(R\times n)}}},\boldsymbol{\sum^{-1}}\in\mathbb{R^{\mathrm{R\times(n\times n)}}},\boldsymbol{W}\in\mathbb{R^{\mathrm{R*(2n+1)\times1}}},\boldsymbol{Z_1}\in\mathbb{{\mathrm{m\times n}}}),...,layer_{D}(\boldsymbol{\widetilde{C}}=[\underline{C},\overline{C}]\in\mathbb{R^{\mathrm{\mathrm{R\times n}}}},\boldsymbol{\sum^{-1}}\in\mathbb{R^{\mathrm{R\times(n\times n)}}},\boldsymbol{W}\in\mathbb{R^{\mathrm{R*(2n+1)\times1}}}))$
		    \FOR{$t=1$ to $N$}
		    \FOR{$j=1$ to $n$}
		    \FOR{$q=1$ to $n$}
		    \STATE \textit{Calculate}: the similarity of two input features  $\gamma(x_{j},x_{q}), j \neq q$ from the input weight vector  $\lambda\in\mathbb{R^{\mathrm{1\times n}}}$ 
		    \ENDFOR
		    \ENDFOR
		    \STATE \textit{Weigh}: the input vector 
		    \IF{$depth=0$}
		    \STATE \textit{Create}: the first layer or base building unit $D=1$
		    \ELSE 
		    \FOR{$d=1$ to $D$}
		    \IF{$d=1$}
		    \STATE \textit{Calculate}: $X_{1}=X$
		    \STATE \textit{Elicit}: the output of each layer
		    \ELSIF{$X_{d}=\lambda X+\alpha Y_{d-1}\boldsymbol{Z_{d-1}}$}
		    \STATE \textit{Elicit}: the output of each layer
		    \ENDIF
		    \ENDFOR
		    \FOR{$i=1$ to $D$}
		    \FOR{$k=1$ to $D$,$i \neq k$}
		    \STATE \textit{Calculate}: similarity of two layers $\gamma(y_{i},y_{k}),i\neq k$
		    \IF{$\gamma(y_{i},y_{k})\leq\delta$}
		    \STATE\textit{Merge}: the layer $i$ and $k$
		    \ENDIF
		    \ENDFOR
		    \ENDFOR
		    \STATE\textit{Undertake}: the drift detection method
		    \IF{Drift}
		    \STATE\textit{Introduce}: a new base building unit/layer
		    \ELSIF{Warning}
		    \STATE\textit{Store}: the data batch into a data buffer to be used for future drift
		    \ELSIF{Stable}
		    \STATE\textit{Update}: the deepest hidden layer
		    \ENDIF
		    \ENDIF
		    \ENDFOR
    \end{algorithmic}
\end{algorithm}

\section{Learning Scheme of DSSCN}
This section elaborates the learning policy of DSSCN including the
fundamental working principle of eSCN as the base building block of
DSSCN. Algorithm 1 displays an overview of DSSCN learning algorithm.
DSSCN working procedure starts from the online feature weighting principle adopting the concept of mutual information. An input feature showing significant redundancy is ruled out from the training process by reducing its corresponding input weight. This mechanism does not compress input dimension but actively selects the best input subset for every training observation by means of dynamic weighting strategy of input features minimizing the influence of poor features. Henceforth, the training procedure of each layer is triggered and the first layer of DSSCN receives the original input features. The output of the first base building unit is combined with random projection matrix to induce the so-called random shift. The shifting factor is applied to the original input attributes feeding the next base building unit. This process proceeds up to the deepest layer. 

DSSCN is fitted with the layer pruning mechanism benefiting from the
notion of mutual information. This concept unveils two base building
units possessing strong mutual information. Such base building blocks
do not offer diverse information to rectify learning performance and 
{is fused into one to relieve the computational complexity.} The
last step concerns with the drift detection scenario which evolves
the structure of DSSCNs. This mechanism controls the depth of DSSCNs
where an extra building unit is added and the structure of DSSCNs is
deepened provided that a concept change is observed. This scenario
is built upon the concept of Hoeffding's bound where the drift detection
mechanism is fully automated and free of user-defined parameters. The conflict
level is determined from a statistically sound method of the Hoeffding's bound. 

\subsection{Online Feature Weighting Mechanism}
DSSCN is equipped by the online feature weighting mechanism which
allocates input weights as the importance of input attributes. This
mechanism is meant to minimize the influence of poor features which
often undermines generalization power without retraining {from scratch due to discontinuity issue.} Note that the use of online
feature weighting mechanism underpins a compact and parsimonious structure
because it rules out inconsequential features from the distance calculation
\cite{pratama2016scaffolding}. In realm of evolving data streams, the feature weighting mechanism has been studied as an alternative of the input pruning method and
is claimed to induce the soft-dimensionality reduction. It
offers more stable feature selection mechanism than its counterpart,
the hard-dimensionality reduction approach, because it adapts to dynamic
of given problems - some features play vital role in different time
periods, thus pruning such features causes information loss. 

The online feature weighting mechanism of DSSCN is inspired by the
work in \cite{pratama2016scaffolding} where the maximal information compression index (MICI)
method \cite{mitra2002unsupervised} is made use to calculate the feature similarity. The
underlying idea is to assign low weight to a feature with strong mutual
information. The MICI method presents an extension of Pearson correlation
index where it is robust against translation and rotation of data
streams. {This method can be directly used for the feature weighting
mechanism without transformation because an input feature attains maximum
correlation if the MICI method returns 0 values.} Suppose that the
linear dependency of two input variables, $x_{1},~x_{2}$ are to be measured,
the MICI is formulated as follows:
\begin{align}
\gamma(x_{1},x_{2})&=\frac{1}{2}(var(x_{1})+var(x_{2})-\nonumber\\&\sqrt{(var(x_{1})+var(x_{2})^{2}-4var(x_{1})var(x_{2})(1-\rho(x_{1},x_{2})^{2})}),
\end{align}
where$\rho(x_{1},x_{2})=\frac{cov(x_{1},x_{2})}{\sqrt{var(x_{1},x_{2})}}$is
the Pearson correlation index. $var(x_{1}),var(x_{2})$ denote the
variance of two features which can be enumerated on the fly. The MICI
method works by estimating maximum information loss when the input
dimension is compressed by ignoring {one of input features.} After the
MICI is obtained, the similarity score of input attributes is inspected
because correlation of input features has to be analyzed for the rest
$n-1$ features $(\gamma(x_{1},x_{2}),...,\gamma(x_{1},x_{n-1}))$.
The similarity score is defined as the average of mutual information
for all input attributes as follows:
\begin{equation}
Score_{j}=\underset{j=l=1,...,n}{mean(\gamma(x_{j},x_{l}))},l\neq j.
\end{equation}
A normalization is required because it is expected that feature contribution
drops in the high-dimensional input attributes - {a feature weight becomes too small. Input weights}, $\lambda_{j}$,
are normalized in respect to {the maximum similarity score} as follows:
\begin{equation}
\lambda_{j}=\frac{Score_{j}}{\underset{j=1,...,n}{max(Score_{j})}}.
\end{equation}
It is worth noting that the input weights function as the connective
weights of the first layer to the hidden layer of eSCN. Since each
building unit receives the same input information added with different random shifts, the input weighting mechanism is committed in the centralized
manner. In other words, the input weights are spread across each building
unit. This is made possible because every layer is injected with the
same input representation and the only difference merely lies in the
random shift factors which vary across different layers. This approach follows a similar input weighting strategy in \cite{pratama2016scaffolding} but here it is implemented under a deep network structure rather than a shallow one.

\subsection{Hidden Layer Pruning Mechanism}
The notion of network layer pruning mechanism is motivated by the
fact that each base building unit should navigate to an improvement
of learning performance. Two strongly correlated base building blocks
can be merged into one thereby alleviating computational and memory
burdens. In realm of deep learning literature, there exist several
attempts for complexity reduction of DNN but most of which are built
upon the concept of regularization where inconsequential layer is
weighted by small regularization factors \cite{OnlineDeepLearning,stochasticdepth,CirculantProjections}.

The network layer pruning mechanism examines the mutual information
among base building units discovering two base building units with
strong mutual information. This strategy is carried out using the
MICI method as with the online feature weighting mechanism and the
only difference from the online feature weighting mechanism exists
in the variables of interest where the output of two base building
units are evaluated as follows:
\begin{equation}
\gamma(y_{i,}y_{k}), i\neq k,
\end{equation}
where $y_{i},y_{k}$ respectively denote the predictive output of $i,k$
building units. Although the MICI method forms a linear correlation
measure, it can be executed more speedily than the non-linear correlation
measure such as the symmetrical uncertainty method by means of the
entropy measure. The entropy measure often necessitates discretization
of input samples or the Parzen density estimation method. Simplification
is commonly applied in the entropy measure where the training data
is assumed to follow normal distribution as shown in the differential
entropy method. Because the maximum linear correlation is attained
provided that $\gamma(y_{i},y_{k})=0$, the layer pruning condition
is formulated as follows:
\begin{equation}
\gamma(y_{i},y_{k})\leq\delta,
\end{equation}
where $\delta$ stands for a user-defined threshold. The layer pruning process is undertaken
by getting rid of one of the two layers or in other words, the layer
pruning process shrinks the network structure.
The layer pruning mechanism enables hard complexity reduction where
the base building unit is permanently forgotten since redundant layer
won't be recalled again in the future. Unlike soft-complexity reduction
\cite{OnlineDeepLearning, CirculantProjections} where base building units are still retained in the memory,
the hard complexity reduction attains great memory simplification
by reducing the depth of network structure. {This mechanism does not result in the catastrophic forgetting problem or instability issue because of the deep stacked network structure of DSSCN}.

\subsection{Hidden Layer Growing Mechanism}
The selection of DSSCN network depth is fully automated from data
streams. The network layer growing mechanism is governed
by a drift detection mechanism which inserts a new base building unit
if a concept change is identified in the data streams. This strategy
is inspired by the fact that a new concept should be embraced by a
new layer without forgetting of previously learned concept. It
retains old knowledge in order for catastrophic forgetting of past
knowledge which might be relevant again under the recurring drift
situation to be avoided. Although the notion of drift detection mechanism
is often integrated in the context of ensemble learning in which the
size of ensemble expands if the drift is detected, this feature can
be also implemented in realm of the deep stacked network since it
is formed by a collection of local learners injected with different level of abstractions of the original data points. Note that although each layer of DSSCN is connected in series,
each layer locally learns data streams with minor interaction with
other layers.

{DSSCN makes use of the Hoeffding's bound drift detection mechanism
classifying data stream into three categories: normal, warning
and drift \cite{drift}.} That is, the normal condition indicates stable concept
which can be handled by simply updating the current network structure.
The update strategy applies only to the last base building unit because of the inter-related nature of deep stacked architecture where one base building unit is influenced by the output of preceding layer or base building unit. This strategy aims to induce different representations of the original training pattern where each hidden layer portrays different perspective of the overall data distribution. It differs from the ensemble configuration where one base building
unit is decoupled from the rest of ensemble members and under this configuration the winning learner should be selected for the fine-tuning phase. The drift phase
pinpoints an uncovered concept which calls for enrichment of existing
structure by integrating a new layer. The warning phase is designed
to cope with the gradual drift since it signals immature drifts requiring
several training observations to be borne out as a concrete drift.
No action is carried out in the warning phase except to prepare existing
structure for a drift case in the future. It implies only accumulation of data points during the transition period, \textit{warning-to-drift}, to be used in the drift case where a new layer is added. It is worth noting that
the gradual drift is more difficult to be tackled than the sudden
drift because the drift can be confirmed at very late stage after the learning
performance has been compromised. The non-weighted moving average
version of the Hoeffding's bound drift detection mechanism \cite{drift} is implemented
in DSSCN because it is more sensitive to the sudden drift than the
weighted moving average version. The statistics of data streams can
be enumerated without any weights as $\widehat{X}_{t}={\textstyle \sum_{t=1}^{N}\frac{x_{t}}{N}}$.

The three conditions of data streams, namely normal, warning and drift
are determined from the Hoeffding test applying two conflict levels:
$\alpha_{W}$ (warning), $\alpha_{D}$(drift) which correspond to
the confidence level of the Hoeffding's statistics as follows:
\begin{equation}\label{conflict}
\varepsilon_{\alpha}=(b-a)\sqrt{\frac{(N-cut)}{2cut(N-cut)}ln(\frac{1}{\alpha})},
\end{equation}
where $[a,b]$ denotes the minimum and maximum of data streams and $\alpha$ is
the significance level where $\alpha_{W}>\alpha_{D}$. $cut$
denotes the cutting point which indicates the inflection point from
one concept to another. Note that the significance level has a clear
statistical interpretation because it also functions as the confidence
level of Hoeffding's bound $1-\alpha$. The drift detection mechanism
partitions a data chunk into three groups $\boldsymbol{Z_{1}}=[x_{1},x_{2},...,x_{cut}],~\boldsymbol{Z_{2}}=[x_{cut+1},x_{cut+2},...,x_{N}]$,~$\boldsymbol{Z_{3}}=[x_{1},x_{2},...,x_{N}]$
where the cutting point corresponds to a switching point signifying
a case of $\hat{Z_{1}}+\varepsilon_{\widehat{Z}_{1}}\geqslant\hat{Z_{3}}+\varepsilon_{\widehat{Z}_{3}}$.
{$\hat{Z}_{1},\hat{Z_{3}}$ are statistics respectively calculated from
two data groups $\boldsymbol{Z_{1}},\boldsymbol{Z_{3}}$.} In other words, the cutting point
refers to a case when a population mean increases. The drift detection
procedure continues by forming the Hoeffding's test which decides
{the status of data streams. The} drift status is returned if the null
hyphotesis is rejected with the size of $\alpha_{D}$, while the warning
status makes use of $\alpha_{W}$ to reject the null hyphotesis. The
null hyphotesis is defined as $H_{0}:E(\hat{Z_{1}})\leq E(\hat{Z_{2}})$ while
its alternative is formulated just as the opposite. The null hyphotesis
is rejected if $\left|\hat{Z}_{1}-\hat{Z}_{2}\right|\geq\varepsilon$ where
$\varepsilon$ is found from {(\ref{conflict})} by applying the specific significance
level $\alpha_{W},\alpha_{D}$. {The hypothesis test} is meant to investigate
the increase of population mean which hints the presence of concept
drift. If the null condition is maintained, existing concept remains valid. Suppose that $dist=|\hat{Z_{1}}-\hat{Z_{2}}|$, the three levels, namely drift, warning and drift, are signalled in respect to the following conditions.
\begin{equation}
    dist\geq \epsilon_{drift}\rightarrow DRIFT
\end{equation}
\begin{equation}
    \epsilon_{warning} \leq dist\leq \epsilon_{drift} \rightarrow WARNING
\end{equation}
\begin{equation}
    ELSE \rightarrow STABLE
\end{equation}
where $\epsilon_{drift},\epsilon_{warning}$ are obtained by substituting the confidence levels $\alpha_D,\alpha_W$ into {(\ref{drift}).}

From Algorithm 1, it is seen that an extra layer is inserted if the
drift status is substantiated by the drift detection mechanism. This
mechanism aims to incorporate a new concept using a new layer while
retaining previous knowledge in preceding layers. The warning phase
does not perform any adaptation mechanism since it presents a waiting
period whether the stable concept turns into the concept drift. Since
the stable status confirms relevance of existing learners for current
data streams, only an update is committed to the last building unit. This approach reflects the nature of deep network structure where each layer is interconnected in terms of random shift. We select to only adjust the last layer because it has the most adjacent relationship to the current training concept. Other layers are left untrained with the current data stream to generate different levels of feature representation and to avoid the catastrophic forgetting problem - the key success of DNNs in the continual environment. This strategy is also designated to overcome the catastrophic forgetting problem

It is evident that the DNN concept is not feasible for small datasets.
A shallow neural network is preferred for {small dataset} since it
achieves a faster convergence due to less free parameters to learn
than the DNNs. As next examples are observed, the performance of shallow
neural networks is stuck at a particular level due to lack of network capacity and is able to be rectified
by introducing extra layers in the neural network structure. In other
words, the success of DNNs's deployment is subject to sample's availability.
The significance level $\alpha$ plays vital role to arrive at the appropriate
network architecture of DNNs since it controls the depth of neural
network. It is not assigned a fixed value here rather constantly adjusted
to adapt to current learning conditions where it grows proportionally
as a factor of the number of data streams. That is, the confidence
level is lowered to allow additional layers or building units to be
integrated. The following setting is formulated to assign the confidence
level:
\begin{equation}
\alpha_{D}=min(1-e^{\frac{-t}{\tau}},\alpha_{min}^{D}),\alpha_{W}=min(1-e^{\frac{-t}{\tau}},\alpha_{min}^{W}) \label{drift},
\end{equation}
where $t$ denotes the number of data streams seen thus far. $\tau$
is a time-constant which can be set as that of the number of time-stamp $N$.
If the number of time-stamp is unknown as practical scenarios, $\tau$
can be assigned a particular value to produce moderate rise of exponential function.
$\alpha_{min}^{D},\alpha_{min}^{W}$ stand for the minimum significance
level of drift and warning phases respectively fixed at 0.09 and 0.1.
This setting is meant to set the confidence level to be above 90\%
which avoids loss of control for the increase of network depth. This
considers the fact that the hidden layer growing strategy is built
upon the drift detection scheme which risks on the loss of detection aptitude
provided that the minimum confidence level is set too low. Referring
to {(11)}, it is assumed that the significance degree mimics the first-order
system dynamic where it exponentially increases as the factor of time
stamp. It is, however, limited to a certain point to avoid too small confidence
level leading to be too greedy in generating new layers. The rationale of drift detection method for layer growing mechanism is elaborated in {Section 7 of this paper.} 
\begin{center}
	\begin{figure}\label{learning architecture of eSCN}
		\begin{centering}
			\includegraphics[scale=0.45]{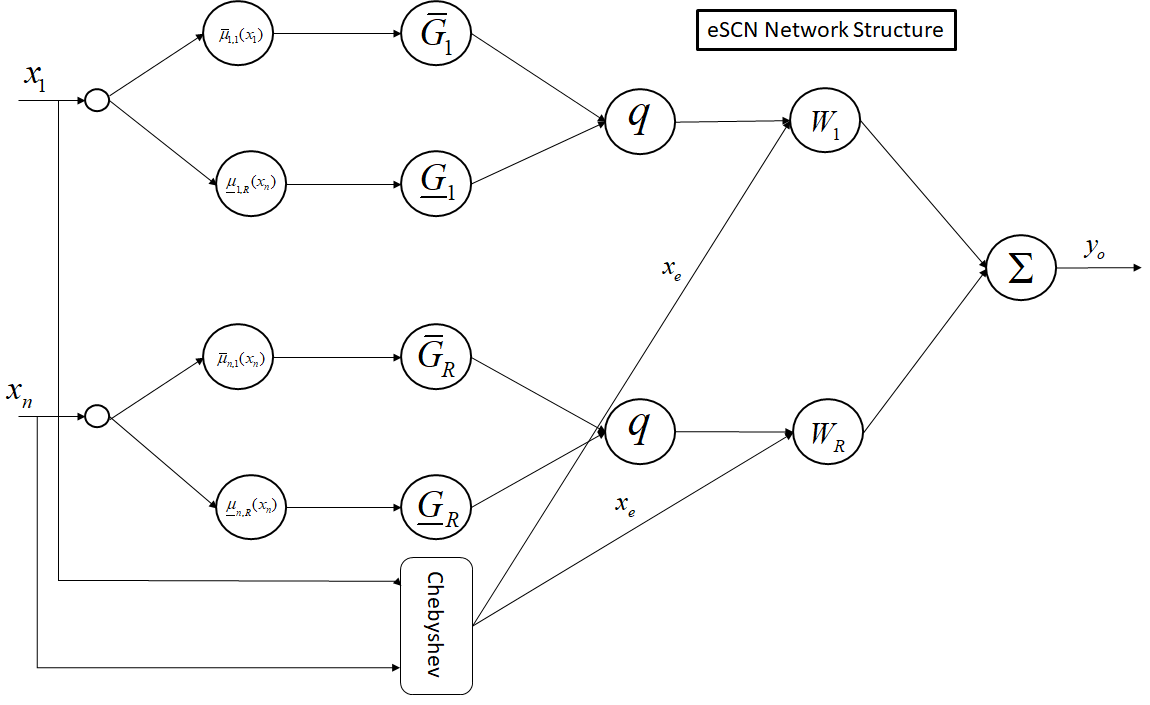}
			\par\end{centering}
		\caption{Learning Architecture of eSCNs}
	\end{figure}
\par\end{center}

\subsection{Evolving Stochastic Configuration Networks}
This sub-section reviews the working principle of eSCN playing a role as a hidden layer of eSCN. eSCN features
a fully open structure in which its hidden node can be automatically
generated and pruned on the fly while following the randomized learning
principle of stochastic configuration network (SCNs). eSCN is a derivation
of eRVFLN \cite{et2rvfln} where the randomized learning foundation is constructed
from the recent development of SCNs. Unlike its predecessor, two learning
modules, online active learning and online feature selection, are
switched off because the online feature selection is carried out in
{the centralized mode} based on the online feature weighting strategy
putting forward the feature redundancy principle (Section 5.1). It
is found that the use of online active learning scenario in each base
building unit is not effective in reducing training samples and labelling
cost since this approach is model-dependent and leads each base building
unit to extract different data subsets - impossible to be carried
out under the deep stacked network configuration. It is supposed
to work more efficiently {in the centralized configuration} which sends
same data samples for each building unit. eSCN learning scenario starts
from scratch and the network structure is self-constructed from
data streams. The contribution of data streams are estimated by a
generalized recursive density estimation principle \cite{angelov2004approach}. The validity
of fuzzy rules are monitored using the type-2 relative mutual information
method and scans for inconsequential rules to be obviated for the
sake of network simplicity. Figure \ref{learning architecture of eSCN} pictorially exhibits the working
principle of eSCN. The learning algorithms of eSCN are detailed as
follows:
\begin{itemize}
    \item 
    Network Structure: eSCNs utilize a combination of the multivariate Gaussian basis function
and {the second order Chebyshev functional expansion block} which
transforms the original input pattern into a set of linearly independent
functions of the entire input attributes \cite{patra2002nonlinear}. The output of Chebyshev
functional expansion block is weighted by the activation degree of
each hidden node generated by the multivariate Gaussian basis function and
the output weights. In other words, the output layer consists of linear
combinations of input features and output weights resulting in the
second order polynomial series. This notion is inspired by the functional-link
neural network structure (FLNN) where it features the so-called enhancement
node combining two components: a direct connection of the input layer
to the output layer and a functional expansion block performing a
non-linear transformation of original input space to a higher dimensional
space \cite{pao1992functional}. The multivariate Gaussian basis function is developed by a non-diagonal covariance matrix featuring inter-correlation between
two input variables. 
Suppose that a tupple $(X_{t},Y_{t})$ streams at time-instant $t-\text{th}$
where $X_{t}\in\Re^{n}$ denotes an input feature of interest and
$Y_{t}\in\Re^{m}$stands for the target variable, the end output of
DSSCN can be expressed as follows:
\begin{equation}
{
y_{o}=\sum_{i=1}^{R}\tilde{G}_{i}(\lambda_t\cdot X_{t}+B_{t})\beta_{i},
}
\end{equation}
where ${\lambda_t\in\Re^{n}}$is the input weight vector at time-instant $t-\text{th}$ obtained from the online feature weighting mechanism in {Section 5.1} and $\tilde{G}$ is the activation function of eSCN defined as the interval-valued multivariate Gaussian function. $B_{t}$ is the network bias set as zero for simplicity, while
$R$ is the number of hidden nodes, $n$ is the number of input dimension
and $m$ is the number of output dimension. $\beta_{i}=x_{e}W_{i}$
where $x_{e}\in\Re^{1\times(2n+1)}$is the extended input vector produced
by the functional expansion block of second order Chebyshev function
while $W_{i}\in\Re^{(2n+1)\times1}$is the output connective weight
vector. The network bias is simply set to zero due to the nature of interval type-2 fuzzy system as realized by eSCN and eT2RVFLN \cite{et2rvfln}. Note that the interval type-2 fuzzy system is a universal approximator \cite{universal_it2f2}. The functional-link strategy in the output layer expands the
degree of freedom (DOF) supposed to improve approximation power of
the output node. Note that the original SCN is developed using the
zero-order output node where each hidden unit is connected with a single connective weight thus leading to $\boldsymbol{W}\in\Re^{R}$. The extended input vector, $x_{e}$, is crafted
from \textbf{red}{a nonlinear mapping of the second order Chebyshev polynomial expansion} written \cite{Chebyshev} as follows:
\begin{equation}
T_{n+1}(x_{j})=2x_{j}T_{n}(x_{j})-T_{n-1}(x_{j}),
\end{equation}
where $T_{0}(x_{j})=1,T_{1}(x_{j})=x_{j},T_{2}(x_{j})=2x_{j}^{2}-1$.
Suppose that the number of input dimension is two, the extended input
vector, $x_{e}$, is set as:
\begin{equation}
x_{e}=[1,T_{1}(x_{1}),T_{2}(x_{1}),T_{1}(x_{2}),T_{2}(x_{2})]=[1,x_{1},2x_{1}^{2}-1,x_{2},2x_{2}^{2}-1].
\end{equation}
The intercept, 1, is inserted in the extended input vector and is
meant to prevent the untypical gradient problem. In addition, it increases
flexibility allowing the extended input vector to go beyond the origin
if all input attributes {are zero}. The functional expansion
block is performed by the Chebyshev polynomial here rather than other
functional-link types such as trigonometric or polynomical functions
since it scatters less number of parameters than trigonometric function
and offers better mapping capability than other polynomial functions
of the same order \cite{Chebyshev}. 

A crisp activation degree is produced
by performing the type reduction mechanism using the $q$ type-reduction mechanism of the interval-valued activation
degree $\widetilde{G}_{i}$ \cite{pratama2016evolving,universal_it2f2} as follows: 
\begin{equation}
\widetilde{G}_{i}=(1-q)\overline{G}_{i}+q\underline{G}_{i},
\end{equation}
where $q\in\Re^{1\times m}$ is a design coefficient which adjusts the influence of upper and lower activation degrees. $\underline{G}_{i},\overline{G}_{i}$ denote
the lower and upper activation functions satisfying the condition
of $\overline{G}_{i}>\underline{G}_{i}$. The upper and lower activation
functions $\underline{G}_{i},\overline{G}_{i}$ are resulted from the
multivariate Gaussian basis function with uncertain means characterized
by the non-diagonal covariance matrix \cite{pratama2016evolving}. This strategy adopts
the notion of interval type-2 fuzzy system presenting the footprint
of uncertainty \cite{pratama2016evolving} which aims to capture possible inaccuracy, imprecision
and uncertainty of data streams due to noisy measurement, noisy data,
operator's error, etc. The $q$ type reduction method \cite{pratama2016evolving} is applied here because it
characterizes a faster operation to transform the inter-valued activation
degrees to its crisp form than the prominent Karnik-Mendel (KM) method \cite{universal_it2f2}
usually involving the iterative process to obtain the right-most and
left-most points. {The interval-valued multivariate Gaussian basis
function, $\widetilde{G}_{i}$, is expressed as follows:}
\begin{equation}
\widetilde{G}_{i}=\exp(-(X-\widetilde{C}_{i}){\boldsymbol\Sigma_{i}^{-1}}(X-\widetilde{C}_{i})^{T}),\quad\widetilde{G}_{i}=[\underline{G}_{i},\overline{G}_{i}],
\end{equation}
where $\widetilde{C}_{i}=[\underline{C}_{i},\overline{C}_{i}]$ denotes
the interval-valued centroid and $\underline{C}_{i}<\overline{C}_{i}$.
$\boldsymbol{\Sigma_{i}^{-1}}$ labels the inverse non-diagonal covariance matrix
whose elements describe inter-relation between two input features.
The interval-valued multivariate Gaussian basis function generates
the non-axis-parallel ellipsoidal cluster in the product space which
has an aptitude to rotate in any direction. Such ellipsoidal cluster
offers flexibility in capturing irregular data distribution which
does not span in the main axes. This property possesses low fuzzy
rule demand which compensates possible increase of network parameters
as a result of the non-diagonal covariance matrix \cite{GENSMARTEFS}.

{The interval-valued activation degree (12)} can be derived further using
the principle of interval type-2 fuzzy system where the expression
of upper and lower activation functions $\underline{G}_{i},\overline{G}_{i}$ can
be obtained. This step, however, calls for a transformation mechanism to adapt to the working formula of interval type-2 Gaussian
function with uncertain means. The transformation strategy finds a one-dimensional representation of high-dimensional Gaussian
function where the center-to-cutting point distance of the non-axis-parallel
ellipsoidal cluster is analyzed. This transformation strategy offers
a fast transformation mechanism because no eigenvalues and eigenvectors
have to be calculated in every observation but it is rather inaccurate
to represent an ellipsoidal cluster with around 45 degrees rotation.
Since the centroid of the ellipsoidal cluster $C_{i}$ remains the
same when projected to one dimensional space, the focus of transformation
strategy is to elicit the radii of high-dimensional multivariate Gaussian
activation $\sigma_{i}$ in the low dimension as follows:
\begin{equation}
\sigma_{i}=\frac{\widetilde{r}_{i}}{\boldsymbol{\sqrt{\Sigma_{i,i}}}}=\widetilde{r}_{i}\sqrt{\boldsymbol{\Sigma_{i,i}^{-1}}},
\end{equation}
where $\widetilde{r}_{i}=\frac{\underline{r}_{i}+\overline{r}_{i}}{2}$
is the Mahalanobis distance of the $i$th rule to an incoming samples
and $\boldsymbol{\Sigma_{i,i}^{-1}}$ is the diagonal elements of inverse covariance
matrix. The same centroid can be applied in the one-dimensional space
without any amendment. Once the low dimensional representation of
multivariate Gaussian function is extracted, the same concept of interval
type-2 fuzzy system can be adopted to quantify the interval-valued
activation degrees as follows:
\begin{equation}
\overline{\mu}_{i,j}=\begin{cases}
N(\underline{c}_{i,j},\sigma_{i,j};x_{j}), & x_{j}<\underline{c}_{i,j}\\
1 & \underline{c}_{i,j} \leq x_j \leq \overline{c}_{i,j}\\
N(\overline{c}_{i,j},\sigma_{i,j};x_{j}) & x_{j}>\overline{c}_{i,j},
\end{cases}
\end{equation}
\begin{equation}
\underline{\mu}_{i,j}=\begin{cases}
N(\overline{c}_{i,j}.\sigma_{i,j};x_{j}) & x_{j}\leq\frac{\overline{c}_{i,j}+\underline{c}_{i,j}}{2}\\
N(\underline{c}_{i,j},\sigma_{i,j};x_{j}) & x_{j}>\frac{\overline{c}_{i,j}+\underline{c}_{i,j}}{2},
\end{cases}
\end{equation}

Once the activation degree per input attribute is elicited using {(18),
(19),} it is combined using the product t-norm operator to induce the
upper and lower activation degrees $\underline{G}_{i}$,$\overline{G}_{i}$ as
follows
\begin{equation}
\underline{G}_{i}=\prod_{j=1}^{n}\underline{\mu}_{i,j},\overline{G}_{i}=\prod_{j=1}^{n}\overline{\mu}_{i,j}.
\end{equation}
The use of the product t-norm operator in {(20)} opens possibility to
apply the gradient-based learning approach compared to the min t-norm
operator. Nevertheless, the flaw of this t-norm operator is apparent
when dealing with a high input dimension {because activation degree decreases as the input variables increase due to the nature of membership function spanning from 0 to 1}. For simplicity, the final output expression of DSSCN {(12)} can be also expressed in one compact form as follows:
\begin{equation}
y_{o}=\frac{(1-q_{0})\sum_{i=1}^{R}\underline{G}_{i}\beta_{i}+q_{0}\sum_{i=1}^{R}\overline{G}_{i}\beta_{i}}{\sum_{i=1}^{R}(\overline{G}_{i}+\underline{G}_{i})},
\end{equation}
where $q_o$ is the design coefficient of the $o-th$ target class performing the type reduction mechanism. The MIMO architecture is implemented here to infer the final predicted class label of eSCN where each rule comprises multiple output weights per each class attribute. That is, the MIMO architecture is supposed to resolve the
class overlapping issue because it takes into account each class specifically
by deploying different sets of output weights per class \cite{pratama2015pclass}. This approach is more robust in handling the class overlapping problem than the popular one versus rest method because it retains the original class structure. The final
predicted class of eSCN is written as follows:
\begin{equation}
y=\underset{1 \leq o\leq m}{max(y_{o})}.
\end{equation}
The MIMO structure starts from encoding of true class label to the
vector form where a component is assigned to ``1''. Suppose that
if the true class label is 2 and the output dimension is 3, the true
class label vector is expressed as $T=[0,1,0]$. This scenario differs
from the direct regression of class label which does not cope with steep class changes. This drawback is caused by the nature of
regression-based classification scheme {inducing} a smooth transition when regressing to the true class label. Figure \ref{Architecture of eSCN} exhibits the network architecture of eSCN.
	\item 
		Hidden Unit Growing Strategy: The hidden node growing procedure of eSCN is akin to that of eRVFLN exploiting the concept of recursive density estimation \cite{angelov2004approach}. It distinguishes itself from the original recursive density estimation method in the use of weighting factor which aims to address the large pair-wise distance problem after observing outliers or remote samples \cite{fuzzypassiveaggressive}. The underlying working principle of this approach is to calculate an accumulated distance of a sample with all other samples seen thus far without keeping all samples in the memory. Moreover, our approach is designed for a non-axis-parallel ellipsoidal cluster whereas the original version only covers the hyper-spherical cluster case. The advantage of this approach is relatively robust against outliers because it takes into account overall spatial proximity information of all samples. The hidden node growing criteria are formulated under two conditions. The first condition is set from the fact where a sample having high summarization power improves generalization potential of hidden nodes. A data sample with the highest density is added to
		be a new hidden node. The second condition is defined to expand the coverage of existing structure which captures non-stationary environments. This setting is realized by incorporating samples with the lowest density which indicates possible concept change. This condition leads to the necessity of a hidden node pruning scenario because such rule may not be supported by the next training samples. Nevertheless, this condition is prone to outlier, thus calling for the hidden node pruning scenario to obviate superfluous hidden nodes which do not play significant role during their lifespan. In addition, the rule replacement condition as known as \textbf{the Condition B} in \cite{angelov2004approach} {is integrated to set a condition for an incoming sample to replace the closest hidden node.} This condition addresses the overlap among hidden nodes and is undertaken by checking the firing strength of the new hidden node candidate (9). The similarity threshold is determined from the chi-square distribution with 5\% to 30\% critical level.  
	\item Hidden Node Pruning Strategy: The hidden node pruning role is warranted
	to attain a compact and parsimonious network structure where it is
	capable of detecting and pruning inactive hidden nodes. This learning
	module prevents overfitting situation which exists due to many clusters
	sitting adjacent to each other. eSCN benefits from the type-2 relative
	mutual information method (T2RMI) which presents an extension of RMI
	method for the type-2 fuzzy system. This method analyzes relevance
	of hidden node in respect to the current data trend by quantifying
	correlation of hidden nodes and target classes. That is, superfluous
	hidden nodes are specified as those having little correlation with
	up-to-date target concept induced by changing data distributions
	or outliers mistakenly inserted as hidden nodes. In the original eT2RVFLN,
	the symmetrical uncertainty method is used to measure the correlation
	of input features and the correlation of hidden units to the target classes. The MICI method can be also exploited as an alternative.
	The symmetrical uncertainty method forms a nonlinear correlation measure
	where it is based on the information gain and entropy analysis. This
	method utilizes the differential entropy concept using the normal
	distribution assumption. 
	\item Parameter Learning Strategy: all network parameters except the output
	weights are randomly initialized using the SCN concept while the output
	weights are updated on the fly using the fuzzily weighted generalized
	recursive least square (FWGRLS) method \cite{pratama2015pclass}. The FWGRLS method can
	be seen as a generalization of FWRLS method \cite{angelov2004approach} where it distinguishes
	itself from its predecessor in the use of weight decay term in the
	cost function of RLS method \cite{rubio1}. This leads to completely different
	tuning formulas but under some simplifications the final update formulas
	differs only in the presence of additional weight decay term. The
	weight decay term limits the outputs weights in small values which
	improves model's generalization. This trait also underpins a compact
	and parsimonious network structure since an inactive hidden node can
	be deactivated by forcing its output weights to small values. The
	use of weight decay term adopts similar idea of generalized recursive
	least square (GRLS) method \cite{GRLS} but the GRLS method is designed under the
	global optimization framework where a global covariance matrix encompassing
	all hidden nodes is constructed. The global learning scenario is considered
	less feasible in the structural learning situation than the local approach
	since changing system structure calls for a resetting of output covariance matrix.
	Furthermore, the local learning scenario has been proven to be more
	robust against noise \cite{EFSSURVEY}. 
	\item Stochastic Configuration Principle: the theory of SCN has
	come into picture to resolve the underlying bottleneck in building randomized learners. The readers must be aware that the universal approximation property of randomized
	approaches do not always hold for the random parameters taken from the uniform distribution over the range of {[}-1,1{]}. It is worth mentioning that SCN starts its training process from scratch with
	no initial structure. The hidden node is incrementally constructed
	by virtue of the hidden node growing procedure. Details of the learning algorithms for building SCN can be read in \cite{SCN} and it is omitted here for simplicity.
	
	Since DSSCN works in the one-pass learning mode which prohibits to revisit
	preceding samples, the SC-II algorithm is implemented. That is, the sliding window has to take at most as the size
	of data chunk and has to be completely discarded after being used. For simplicity, the size of sliding window is fixed
	as the size of data chunk. Table 2 outlines the pseudocode of
	SCN-based randomization strategy. The concept of SCN is applied when adding a new hidden node or when satisfying the hidden node growing condition determined using an extension of the recursive density estimation method \cite{angelov2004approach}. It randomly select the inverse covariance matrix of the multivariate Gaussian function from a dynamic scope while the center of the multivariate Gaussian hidden node is set as an incoming data sample passing the hidden node growing condition. The output weight vector of a new hidden node is set as that of the winning node. The underlying reason is its closest
	proximity to the new hidden node, thus it is expected to characterize
	similar approximation curve. The output covariance matrix of the new
	hidden node in the FWGRLS method is assigned as $\boldsymbol{\Omega_{R+1}}=\omega I$
	where $\omega$ is a positive large constant. It is well-known that
	the convergence of FWGRLS method is assured when the new output covariance matrix is crafted as a large positive definite matrix. This setting addresses the fact that a newly initialized rule still has poor approximation power, {thereby calling for a large correction factor using a large positive definite matrix $\boldsymbol{\Omega_{R+1}}$ \cite{olivernelles}.} It is worth noting that the sliding window is applied to calculate the constraint parameter and the pseudo-inversion method is excluded from the learning procedure of eSCN. The parameter adjustment scenario is done using the FWGRLS method to expedite the training process. Moreover, the convergence of SCN has been theoretically proven \cite{SCN} and the solution of SCN cannot be seen as an ad-hoc solution. SCN produces random parameters that maximizes the robustness variable $\zeta$ from a dynamically generated scope $[-\xi,\xi]$. Note that the predefined constants, $T_{max},\xi_{min},\xi_{max}$, only determines the population of search space or resolution of SCN. The higher the value of $T_{max}$ the higher the number of candidates of random variables is while the increase of $\xi_{min},\xi_{max}$ also increases the possible range of random variables. Table \ref{tab:Network Parameters} displays the network parameters of DSSCN including the number of dimension, the initialization strategy and the learning strategy. 
\end{itemize}
\begin{center}
	\begin{figure}\label{Architecture of eSCN}
		\begin{centering}
			\includegraphics[scale=0.5]{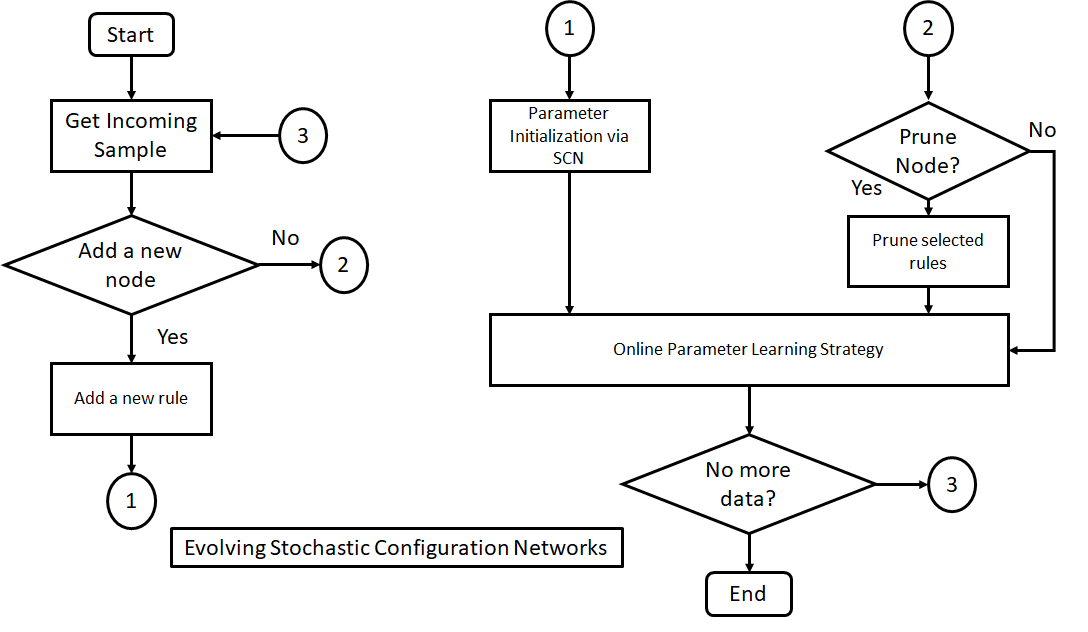}
			\par\end{centering}
		\caption{{Evolving Stochastic Configuration Networks}}
	\end{figure}
	\par\end{center} 
	
\begin{algorithm}
    \caption{{Pseudocode of SCN-based Initialization Strategy}}
	\label{SCN-based Initialization Strategy}
	\begin{algorithmic}[1]
	    \STATE \textit{Input}: Data Chunk $(X,T)\in\mathbb{R^{\mathrm{N\times(n+m)}}}$,\\the
		number of hidden nodes at the current time step $R$, the residual error $\boldsymbol{e_{R}}(X)\in\mathbb{R^{\mathrm{N\times m}}}$, \\the interval-valued activation degrees $\widetilde{\boldsymbol{G}}_{1:R}(X)\in\mathbb{R^{\mathrm{N\times R}}}$the user defined parameters: \\the maximum stochastic configuration $M_{max}$, a set of positive scalars $\mathrm{T=\left\{ \xi_{min},\Delta\xi,\xi_{max}\right\} }$and$0<r<1$.
		\STATE \textit{Output}: Parameters of New Hidden Node $\widetilde{C}_{R+1}=[\underline{C}_{R+1},\overline{C}_{R+1}]\in\mathbb{R^\mathrm{R\times n}}\boldsymbol{\sum_{R+1}}^{-1}\in\mathbb{R^{\mathrm{n\times n}}},W\in\mathbb{R^{\mathrm{R*(2n+1)\times1}}},\boldsymbol{\Omega}\in\mathbb{R^{\mathrm{(2n+1)\times(2n+1)}}}$.
 		\IF{A New Hidden Node is Added}
		\STATE Step 1: Allocate The Center of Activation Function \textbf{$\underline{C}_{R+1}=X_n-\delta,\overline{C}_{R+1}=X_n+\delta$} While Using The SCN Concept for Inverse Covariance Matrix \textbf{$\sum_{R+1}^{-1}$}:
		\STATE \textit{Create}: two empty sets $\Phi,\Psi$
		\FOR{$\xi\in\mathrm{T}$}
		\FOR{$k=1$ to $T_{max}$}
		\STATE \textit{Assign}: the inverse covariance matrix in the range of $[-\xi,\xi]$ randomly
		\STATE \textit{Calculate}: the robustness variable $\zeta_{R+1,o}$ and $\mu_{R+1}=\frac{(1-r)}{(R+1)}$
		\IF{$min(\zeta_{R+1,1},...,\zeta_{R+1,m})\geq0$}
		\STATE \textit{Save}: random parameters $\sum_{R+1}^{-1}$ in the temporary matrix $\Psi$ an $\zeta_{R+1}=\sum_{o=1}^{m}\zeta_{R+1,o}$ in $\Phi$, respectively
		\ENDIF
		\ENDFOR
		\IF{$\Psi$ is not empty}
		\STATE \textit{Choose}: random parameters $\boldsymbol{\sum}_{R+1}^{-1}$maximizing $\zeta_{R+1}$ in $\Phi$ as parameters of new hidden node
		\STATE \textbf{Break}
		\ELSE
		\STATE \textit{Choose}: $\tau\in(0,1-r)$ randomly
		\STATE \textit{Update}: $r=r+1$
		\ENDIF
		\ENDFOR
		\STATE{Step 2: Allocate The Output Weight Parameters}:
		\STATE \textit{Allocate}: the output weight vector and output covariance matrix of	the new hidden node respectively as $W_{R+1}=W_{win},\boldsymbol{\Omega}_{R+1}=\omega I$
		\ENDIF
	\end{algorithmic}
\end{algorithm}

\begin{table}[t]
		\caption{Network Parameters of DSSCN}
		\begin{centering}
		\label{tab:Network Parameters2}
		\scalebox{0.8}{
		\begin{tabular}{cccl}
		\toprule
		\textbf{Parameters} & \textbf{Dimension} & \textbf{Initialization} & Tuning Strategy\tabularnewline
		\midrule
		$\widetilde{C}_i\in[\underline{C}_i,\overline{C}_i]$ & $1\times n$ & $\underline{C}_i=X_n-\delta,\overline{C}_i=\overline{C}_i=X_n+\delta$ & N/A \tabularnewline
		$\boldsymbol{\sum_{i}^{-1}}$ & $n \times n$ & SCN-based initialization strategy & N/A \tabularnewline
		$q$ & $1 \times m$ & user-defined & N/A \tabularnewline
		$\alpha$ & $1 \times n$ & initialized at 1 & the online input weighting scenario - Section 6.1 \tabularnewline
		$\boldsymbol{W}_i$ & $(2n+1) \times m$ & $W_{win}$ & the FWGRLS method \tabularnewline
		$\boldsymbol{\Omega}_i$ & $(2n+1) \times (2n+1)$ & $\omega \boldsymbol{I}$ & the FWGRLS method \tabularnewline
		\bottomrule
		\end{tabular}}
		\par\end{centering}
\end{table}

\section{Performance Evaluation}
This section aims to numerically validate the efficacy of DSSCN by
means of thorough simulations utilizing six synthetic and real-world
data streams and comparisons with prominent data stream analytics
methods in the literature. Simulations are undertaken under MATLAB
environment of an Intel (R) Core i5-6600 CPU @ 3.3 GHZ with 8 GB of
RAM and the MATLAB code of DSSCN is made publicly available in \footnote{\url{https://www.researchgate.net/profile/Mahardhika_Pratama}}.
Three popular artificial data streams, namely SEA \cite{SEA}, SUSY \cite{SUSY}
and Hyperplane \cite{MOA}, are used to test the performance of DSSCN, while
DSSCN learning performance is also examined with two real-world data
streams: electricity pricing valuation and weather prediction
\cite{DitzlerImbalanced}. Another data stream problem is picked up from our indoor RFID localization problem in the manufacturing shopfloor where the underlying goal is
to locate the position of raw materials in the production line.
DSSCN is also compared with data stream algorithms: pENsemble \cite{pENsemble}, Learn++.NSE \cite{Leanr++NSE}, Learn++.CDS \cite{DitzlerImbalanced}, pClass \cite{pratama2015pclass}, eT2Class \cite{pratama2016evolving}. Moreover,
the advantage of DSSCN is indicated by comparison with eSCN with the
absence of deep stacked network architecture. The periodic hold-out process
is followed as our simulation protocol \cite{EFSSURVEY}. That is, each data
stream is divided into two parts where the first part is set for model
updates, while the second part is exploited to assess the model's
generalization. {In addition, Section 6.7 reports our simulation under the prequential test-then-train procedure to show that DSSCN works equally well in two popular evaluation procedure of data stream algorithms.}

\subsection{SEA Dataset}
The SEA problem is a popular problem containing 100 K artificially
generated data samples \cite{SEA}. This problem is well-known for its
abrupt drift component induced by three dramatic changes of the class boundary $\theta=4\rightarrow7\rightarrow4\rightarrow7$.
That is, this problem characterizes a binary classification problem
where data samples are classified as the class 1 if it falls below
the threshold $f_{1}+f_{2}<\theta$ while class 2 is returned if a
sample is higher than the threshold $f_{1}+f_{2}>\theta$. Moreover,
the modified version of Ditzler and Polikar \cite{DitzlerImbalanced} is used in our numerical
study where it differs itself from the original version in the class
imbalanced and recurring drift properties with 5 to 25\% minority
class proportion. Data are drawn from the uniform distribution in
the range of {[}0,10{]} and the prediction is underpinned by three
input attributes but the third input attribute is just a noise.
Data stream environment is generated by batches of 1000 data samples
with 200 time stamps and the concept drift takes place at every 50
time stamp. {Figure \ref{SEAdata}(a)-(d)} pictorially exhibits the trace of classification
rate, hidden node, hidden layer and input weights. Table 3 tabulates
consolidated numerical results of benchmarked algorithms. 
\begin{table}[t]
		\caption{SEA Dataset}
		\begin{centering}
		\label{tab:SEA}
		\scalebox{0.8}{
		\begin{tabular}{lrrrrcl}
				\toprule
				\textbf{Model} & \textbf{Classification Rate} & \textbf{Node} &\textbf{Layer/Local Model} &\textbf{Runtime} &\textbf{Input} &\textbf{Architecture} \tabularnewline
				\midrule
				DSSCN & 0.96$\pm$0.02 &7.64$\pm$4.5 & 2.23$\pm$1.18 & 0.82$\pm$0.14 & 3 & Deep \tabularnewline
				eSCN &	0.95$\pm$0.02&\textbf{1}&1&	0.46$\pm$0.27&	3	&Shallow\tabularnewline
				Learn++NSE& 0.93$\pm$0.02&	N/A&	200&	1804.2&	3&	Ensemble\tabularnewline
				Learn++CDS&	0.93$\pm$0.02&N/A&	200	&2261.1	&3&	Ensemble\tabularnewline
				pENsemble&	\textbf{0.97$\pm$0.02}&	4.06$\pm$1.8	&\textbf{2.03$\pm$0.3}	&0.89$\pm$0.12&	3&	Ensemble\tabularnewline
				pClass&	0.89$\pm$0.1&	6.6$\pm$4.2	&1	&0.42$\pm$0.3&	3&	Shallow\tabularnewline
				eT2Class&	0.88$\pm$0.23&	1.5$\pm$0.5	&1	&\textbf{0.34$\pm$0.11}&	3&	Shallow\tabularnewline
				\bottomrule 
		\end{tabular}}
		\par\end{centering}
\end{table}

The efficacy of DSSCN is evident in Table 3 where DSSCN delivers accurate prediction beating other algorithms except pENsemble. This result, however, should be carefully understood because of different base learners between DSSCN and pENsemble. pENsemble utilizes pClass as the base learner which fully learns all network parameters whereas eSCN is deployed in the DSSCN. That is, eSCN is designed following a randomized approach presenting rough estimation of the true solution. {Figure \ref{SEAdata}(a)}
exhibits that DSSCN responds timely to concept drift where a new layer is introduced provided that a drift is identified. Furthermore, this fact also confirms that addition of new layer can be applied to cope with concept drift. A new layer characterizes recent observations and is placed at the top layer of deep stacked structure controlling the final output of the DSSCN. In addition, the layer pruning mechanism functions to coalesce redundant layers, thereby relieving the model's complexity. Adaptive feature weighting mechanism is exemplified in {Figure \ref{SEAdata}(d)} where an input feature showing high mutual information is
assigned with a low weight to minimize its influence to the training process. Another interesting observation is found in the scope of random parameters. Hidden node parameters are randomly sampled from the range [-0.5,0.5], [-1,1], [-2,2] and [-3,3] automatically found in the training process. These ranges are fully data-driven where the SCN-based initialization strategy is executed whenever a new rule is created.
\begin{table}[t]
		\caption{Weather Dataset}
		\begin{centering}
		\label{tab:Weather}
		\scalebox{0.8}{
		\begin{tabular}{lrrrrcl}
				\toprule
				\textbf{Model} & \textbf{Classification Rate} & \textbf{Node} &\textbf{Layer/Local Model} &\textbf{Runtime} &\textbf{Input} &\textbf{Architecture} \tabularnewline
				\midrule
				DSSCN & \textbf{0.81$\pm$0.02} &\textbf{1.7$\pm$0.7} & 1.7$\pm$0.7 & \textbf{1.23$\pm$0.29} & 8 & Deep \tabularnewline
				eSCN &	0.69$\pm$0.03&15.5$\pm$3.4&1&	6.51$\pm$2.29&	8	&Shallow \tabularnewline
				Learn++NSE& 0.75$\pm$0.03&	N/A&	10&	184.4&	8&	Ensemble \tabularnewline
				Learn++CDS&	0.73$\pm$0.02&N/A&	10	&9.98	&8&	Ensemble \tabularnewline
			    pENsemble&	0.78$\pm$0.02&	3$\pm$1.05	&\textbf{1.5$\pm$0.5}	&1.4$\pm$0.06&	8&	Ensemble \tabularnewline
				pClass&	0.8$\pm$0.04&	2.3$\pm$0.5	&1	&1.8$\pm$0.22&	8&	Shallow \tabularnewline
				eT2Class&	0.8$\pm$0.03&	2.3$\pm$0.3	&1	&1.8$\pm$0.1&	8&	Shallow\tabularnewline
				\bottomrule 
		\end{tabular}}
		\par\end{centering}
\end{table}

\begin{center}
	\begin{figure}[t]\label{SEAdata}
		\begin{centering}
			\includegraphics[scale=0.5]{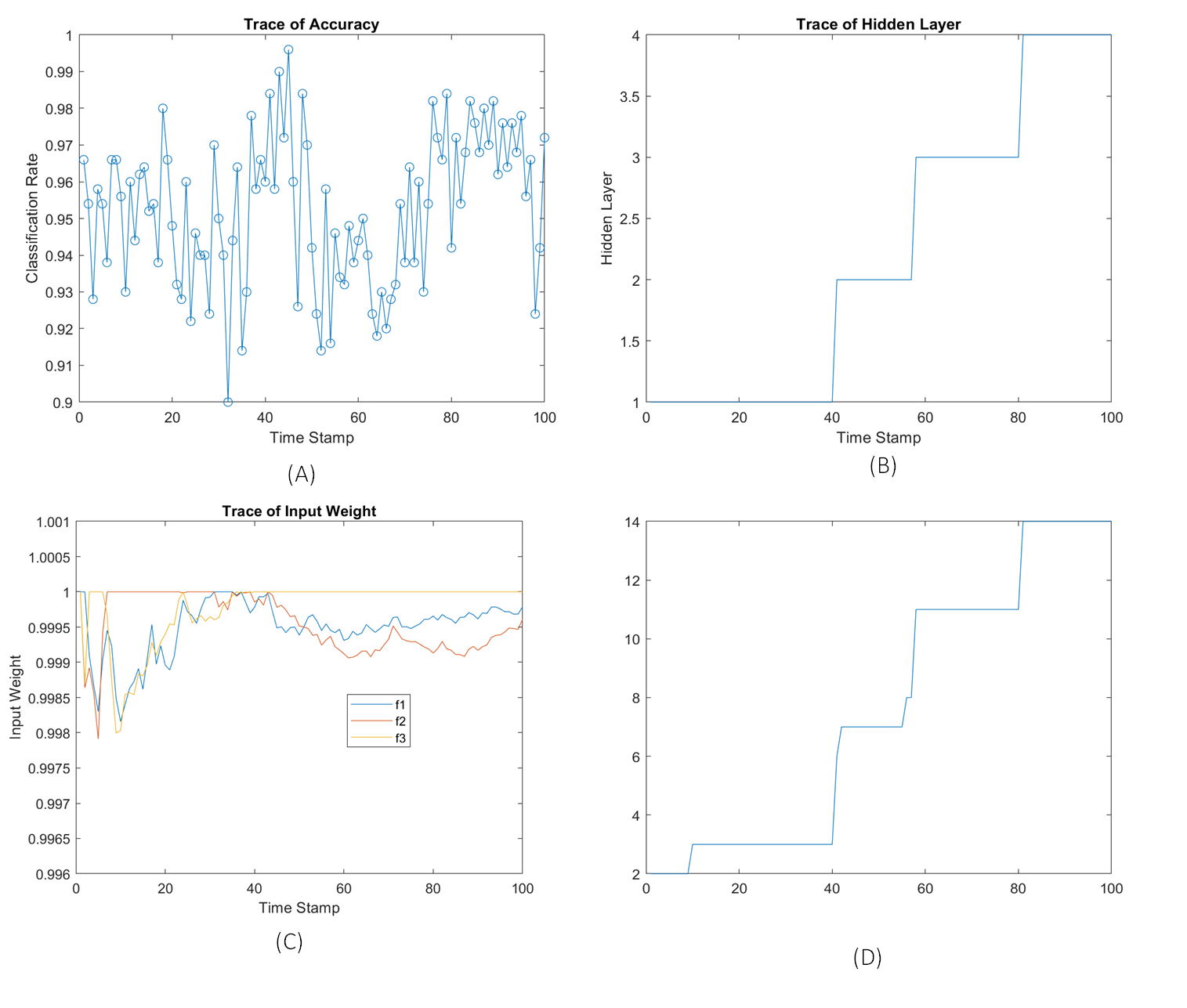}
			\par\end{centering}
		\caption{SEA data}
	\end{figure}
	\par\end{center} 

\subsection{Weather Dataset}
The weather dataset is exploited to examine the efficacy of DSSCN
because it features recurring drift which uniquely refers to reappearance of previously seen data distribution due to {weather seasonal changes.} This dataset is a subset of NOAA database comprising daily weather data from different stations around the world. Among hundreds of weather stations, daily records of weather conditions from Offutt Air Force Base in Bellevue, Nebraska is chosen because it possesses the most
complete recording for over 50 years which allows to portray not only ordinary seasonal change but also long-term climate change \cite{DitzlerImbalanced}. The underlying goal of this problem is to perform a binary classification of one-step-ahead
weather prediction whether rain would occur on the next day with the use of eight input attributes. A total of 60000 data points are involved in our simulation where the average numerical results over ten time stamps are reported in Table 4. {Figure \ref{weatherdata}(a)} exhibits the evolution
of DSSCN classification rates and {Figure \ref{weatherdata}(b)} depicts the evolution of hidden nodes, while the evolution of hidden layer is shown in Figure {\ref{weatherdata}(c)} and the trace of input weights is visualized in {Figure \ref{weatherdata}(d)}.

Referring to Table 4, DSSCN outperforms other benchmarked algorithms in terms of predictive accuracy and hidden layers/local experts. It is observed that DSSCN indicates significant performance improvement over eSCN using a shallow network structure. Compared to pENsemble - an evolving ensemble learner, DSSCN produces slightly more inferior predictive accuracy than pENsemble but involves lower number of local experts. DSSCN also undergoes the fastest training speed than other algorithms. This fact can be explained from the update strategy of DSSCN where only the last hidden layer {experiences the tuning phase.} In addition, eSCN encompasses all data samples of the data chunk during construction of new base building unit to discover
the best scope of hidden node parameters. 
{Figure \ref{weatherdata}(b) and \ref{weatherdata}(c)} confirms the growing and pruning property of DSSCN
in which it is not only capable of self-organizing its hidden nodes but also its hidden layers with addition and removal of base building units. As portrayed in {Figure \ref{weatherdata}(d)}, the online feature weighting mechanism controls the influence of input features where it allows an input feature to pick up again in the future. The SCN-based intialization strategy produces the scope of [-100,100] applied to generate new hidden node parameters. It is worth mentioning that the SCN-based initialization strategy is executed in the main training process of eSCN. 
\begin{center}
	\begin{figure}[t]\label{weatherdata}
		\begin{centering}
			\includegraphics[scale=0.5]{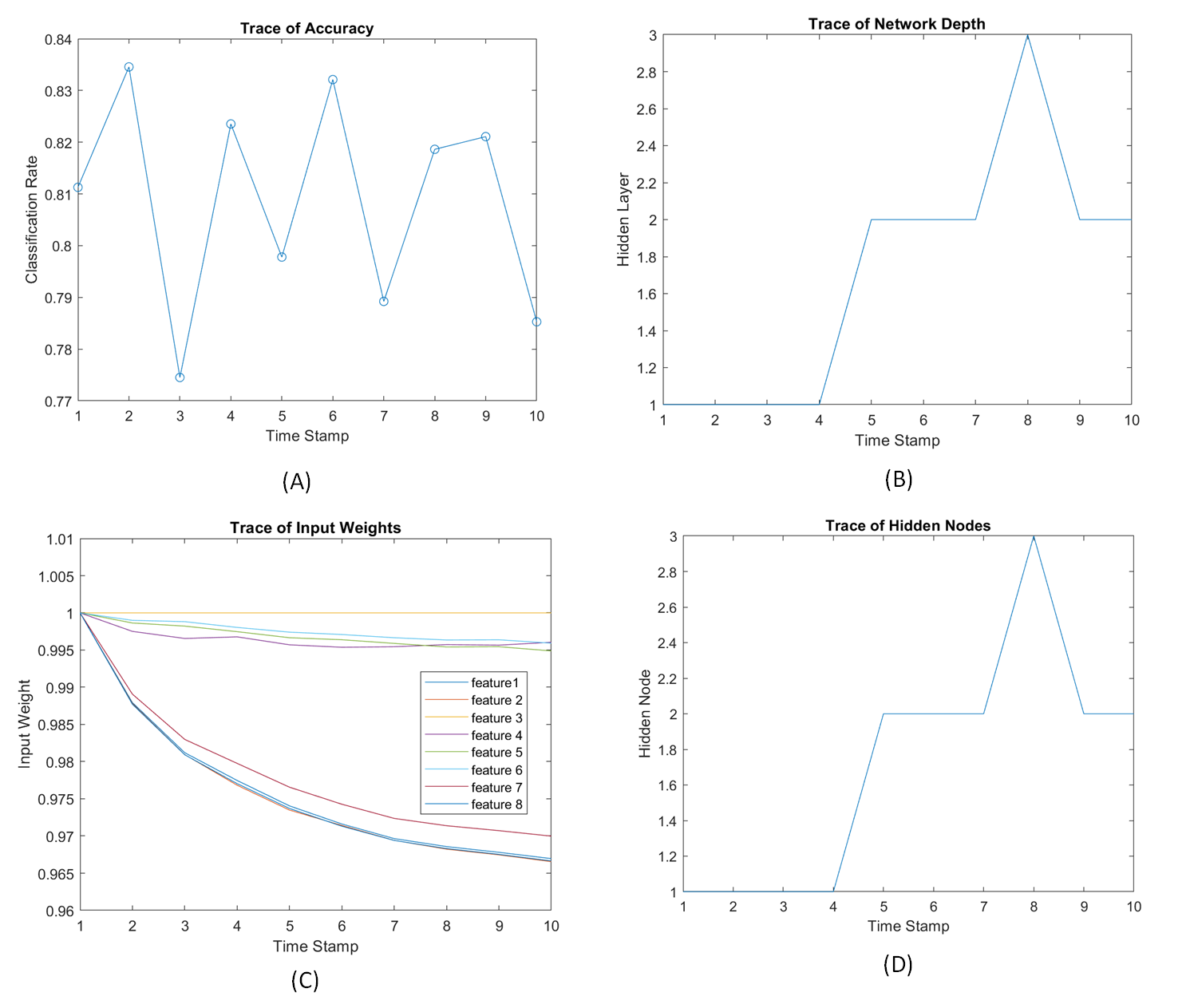}
			\par\end{centering}
		\caption{weather data}
	\end{figure}
	\par\end{center} 

\subsection{Electricity Pricing Dataset}
The electricity pricing dataset describes prediction of electricity
demand fluctuation in the state of New South Wales (NSW), Australia
where the main objective is to forecast whether the electricity price
of NSW in the next 24 hours will be higher than that of Victoria.
{Eight input attributes
are put forward to guide the electricity demand prediction.} Furthermore,
the modified electricity pricing dataset in \cite{DitzlerImbalanced} is used to simulate
the class imbalance problem with the imbalance ratio of 1:18 by performing
undersampling to the minority class. It is worth mentioning that this
problem features the non-stationary characteristic as a result of
dynamic market condition and economic activities. Numerical results
of all consolidated algorithms are reported in Table 5. 
\begin{table}[t]
		\caption{Electricity Pricing Dataset}
		\begin{centering}
		\label{tab:Electricity Pricing}
		\scalebox{0.8}{
		\begin{tabular}{lrrrrcl}
				\toprule
				\textbf{Model} & \textbf{Classification Rate} & \textbf{Node} &\textbf{Layer/Local Model} &\textbf{Runtime} &\textbf{Input} &\textbf{Architecture} \tabularnewline
				\midrule
				DSSCN & 0.734$\pm$0.16 &3.83$\pm$1.82 & 1.4$\pm$0.49 & 0.2$\pm$0.07 & 8 & Deep \tabularnewline
				eSCN &	0.73$\pm$0.17&\textbf{1}&1&	\textbf{0.04$\pm$0.006}&	8	&Shallow\tabularnewline
				Learn++NSE& 0.69$\pm$0.08&	N/A&	119&	211.2&	4&	Ensemble\tabularnewline
				Learn++CDS&	0.69$\pm$0.08&N/A&	119	&211.2	&4&	Ensemble\tabularnewline
				pENsemble&	\textbf{0.75$\pm$0.16}&	1.01$\pm$0.07	&\textbf{1.01$\pm$0.07}	&1.4$\pm$0.06&	8&	Ensemble\tabularnewline
				pClass&	0.68$\pm$0.1&	3.5$\pm$2.4	&1	&7.1$\pm$4.4&	8&	Shallow\tabularnewline
				eT2Class&	0.72$\pm$0.17&	4.6$\pm$1.3	&1	&0.3$\pm$0.08&	8&	Shallow\tabularnewline
				\bottomrule 
		\end{tabular}}
		\par\end{centering}
\end{table}

DSSCN clearly delivers promising performance where it is only inferior
to pENsemble in terms of accuracy by about 1 \%. Compared to its shallow counterpart,
eSCN, the use of a deep network structure is capable of delivering
improvement of model's generalization. The unique trait of DSSCN is
viewed in the incremental construction of deep network structure where
not only hidden nodes can be automatically grown and pruned but also
hidden layer is fully self-organized. This elastic structure
adapts to concept drifts because concept drift can be identified and
reacted in accordance with its severity and rate. The adaptive scope selection property of DSSCN is substantiated by the fact that [-3,3] is selected during the training process.

\subsection{Susy Dataset}

Susy problem actualizes a big streaming data problem consisting of
five millions records allowing algorithmic validation in the lifelong
learning fashion. This problem is a binary classification problem
produced by Monte Carlo simulations and describes classification of
signal process that generates supersymmetric particles \cite{SUSY}. The classification
problem is guided with the use of 18 input attributes where the first
eight input attributes present the kinematic properties while the
remainders are simply the function of the first eight input features.
All benchmarked algorithms are run through 10000 data batches where
each batch carries 500 data samples. 400 data points are reserved
for model updates while the rest of 100 samples are exploited for
validation. The training samples are collected from the first 4.5
millions of data samples while the testing data are sampled from the
next 500 K data points. Numerical results of consolidated algorithms
are summarized in Table 6. {Figure \ref{SUSYdata}(a) visualizes the trace of classification
rates and Figure \ref{SUSYdata}(b) exhibits the trace of hidden nodes, while the
trace of hidden layers and feature weights are displayed in Figure \ref{SUSYdata}(c)
and Figure \ref{SUSYdata}(d) respectively.}
\begin{table}[t]
		\caption{Susy Dataset}
		\begin{centering}
		\label{tab:Susy}
		\scalebox{0.8}{
		\begin{tabular}{lrrrrcl}
				\toprule
				\textbf{Model} & \textbf{Classification Rate} & \textbf{Node} &\textbf{Layer/Local Model} &\textbf{Runtime} &\textbf{Input} &\textbf{Architecture} \tabularnewline
				\midrule
				DSSCN & \textbf{0.78$\pm$0.03} &10.73$\pm$3.4 & 10.73$\pm$3.4 & \textbf{0.34$\pm$0.03} & 18 & Deep \tabularnewline
				eSCN &	0.76$\pm$0.04 & 18.9$\pm$11.92 & 1 &	18.3$\pm$20.08 & 18	&Shallow\tabularnewline
				pENsemble&	0.76$\pm$0.04&	\textbf{1.6$\pm$0.7}	&\textbf{2.84$\pm$1.4}	&0.6$\pm$0.28&	18&	Ensemble\tabularnewline
				pClass&	0.73$\pm$0.06&	1.96$\pm$0.26	&1	&0.79$\pm$0.3&	18&	Shallow\tabularnewline
				eT2Class&	0.74$\pm$0.06&	5.2$\pm$1.5	&1	&2.42$\pm$0.6&	18&	Shallow\tabularnewline
				\bottomrule 
		\end{tabular}}
		\par\end{centering}
\end{table}

From Table 6, DSSCN delivers the highest classification rate. It is
shown that DSSCN's deep hierarchical network and learning modules
has led to 2\% improvement of predictive accuracy of eSCN relying
on a shallow network architecture. DSSCN also outperforms other benchmarked algorithms in term of execution time. This fact differs from the ensemble learner where one can
simply adjust the most relevant ensemble member because each of them
is loosely coupled. {Figure \ref{SUSYdata}(b)} also depicts the incremental constructurion
of deep neural network where new layer can be incorporated on demand
while redundant layers can be merged. It is observed that DSSCN actualizes
dynamic confidence levels of drift detector where initially the confidence
level is set high but gradually reduces by means of the exponentially
decaying function {(11)}. This setting makes possible to increase the
depth of neural network more intensively as more training samples
become available. It is well-known that the success of deep neural
network is subject to sample's availability. Because the drift detection
method is applied to grow the network structure, the decrease of confidence
level is capped at 90\% to prevent introduction of hidden layers during
the drift-free state. Learn++NSE and Learn++CDS are excluded from
this problem because they were terminated before obtaining numerical
results although they have been executed for three days. the scope of [-100,100] is automatically picked up by the SCN-based intialization strategy in this numerical example.
\begin{center}
	\begin{figure}[t]\label{SUSYdata}
		\begin{centering}
			\includegraphics[scale=0.6]{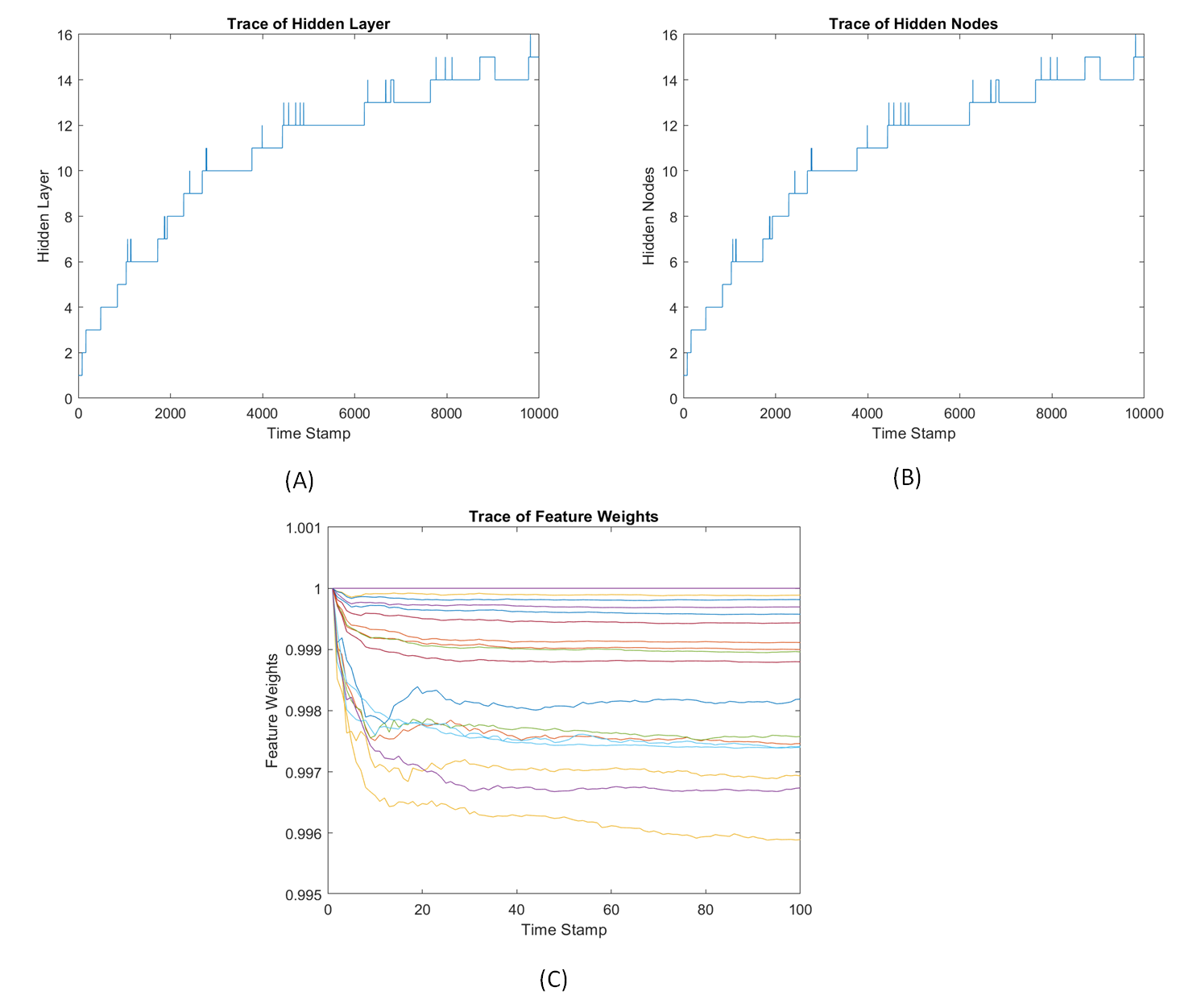}
			\par\end{centering}
		\caption{SUSY data}
	\end{figure}
	\par\end{center} 

\subsection{Hyperplane Dataset}
The hyperplane problem presents a binary classification problem clasifying
a $d$-dimensional data point into two classes with respect to the position
of a random hyperplane. This problem is taken from the Massive Online
Analysis (MOA) \cite{MOA} - an open source software for data stream analytics.
The prominent characteristic of this problem is found in the gradual
drift where initially data samples are drawn from a single data distribution
with probability one. The transition period takes place by the introduction
of the second distribution which slowly interferes data generarion
process before completely replaces the first data distribution. A
data sample is assigned as class 1 if $\sum_{i=1}^{d}x_{i}w_{i}>w_{0}$ is
satisfied whereas class 2 is returned if the condition is violated.
This dataset carries 120 K samples equally partitioned into 100 data
batches without changing data order to simulate data stream environments.
Each data stream comprises 1200 samples where the first 1000 samples
are injected for the training process while the rest is reserved for
testing samples. The concept drift comes into picture after the 40th
time stamp. Table 7 sums up numerical results of all consolidated
algorithms.
\begin{table}[t]
		\caption{Hyperplane Dataset}
		\begin{centering}
		\label{tab:Hyperplane}
		\scalebox{0.8}{
		\begin{tabular}{lrrrrcl}
				\toprule
				\textbf{Model} & \textbf{Classification Rate} & \textbf{Node} &\textbf{Layer/Local Model} &\textbf{Runtime} &\textbf{Input} &\textbf{Architecture} \tabularnewline
				\midrule
				DSSCN & \textbf{0.92$\pm$0.02} &9.3$\pm$1.98 & 1.89$\pm$0.37 & \textbf{0.54$\pm$0.13} & 2 & Deep \tabularnewline
				eSCN &	\textbf{0.92$\pm$0.02}&4.21$\pm$0.54&1&	0.82$\pm$0.15 &	2	&Shallow\tabularnewline
				Learn++NSE& 0.91$\pm$0.02&	N/A&	100&	926.04&	2&	Ensemble\tabularnewline
				Learn++CDS&	0.9$\pm$0.0&N/A&	100	&2125.5	&2&	Ensemble\tabularnewline
				pENsemble&	\textbf{0.92$\pm$0.02}&	2.7$\pm$0.7	&\textbf{1.87$\pm$0.34}	&0.7$\pm$0.23&	2&	Ensemble\tabularnewline
				pClass&	0.91$\pm$0.02&	3.8$\pm$1.7	&1	&2.7$\pm$1.4&	2&	Shallow\tabularnewline
				eT2Class&	0.89$\pm$0.1&	\textbf{2.04$\pm$0.2}	&1	&2.5$\pm$1.5&	2&	Shallow\tabularnewline
				\bottomrule 
		\end{tabular}}
		\par\end{centering}
\end{table}

DSSCN delivers the most encouraging performance in terms of accuracy
and network complexity. Compared to Learn++ family, DSSCN offers more
parsimonious network structure because a new data stream does not
necessarily cause addition of a new local expert due to the use of
drift detection mechanism. It is also observed that DSSCN's runtime
is comparable to that shallow neural network variants because only the last layer is fine-tuned. It is also reported that the scope of random parameters are obtained from the ranges of [-1,1],[-1.5,1.5],[-2,2],[-3,3] confirming the adaptive scope selection of DSSCN.

\begin{center}
	\begin{figure}\label{smartrack}
		\begin{centering}
			\includegraphics[scale=0.35]{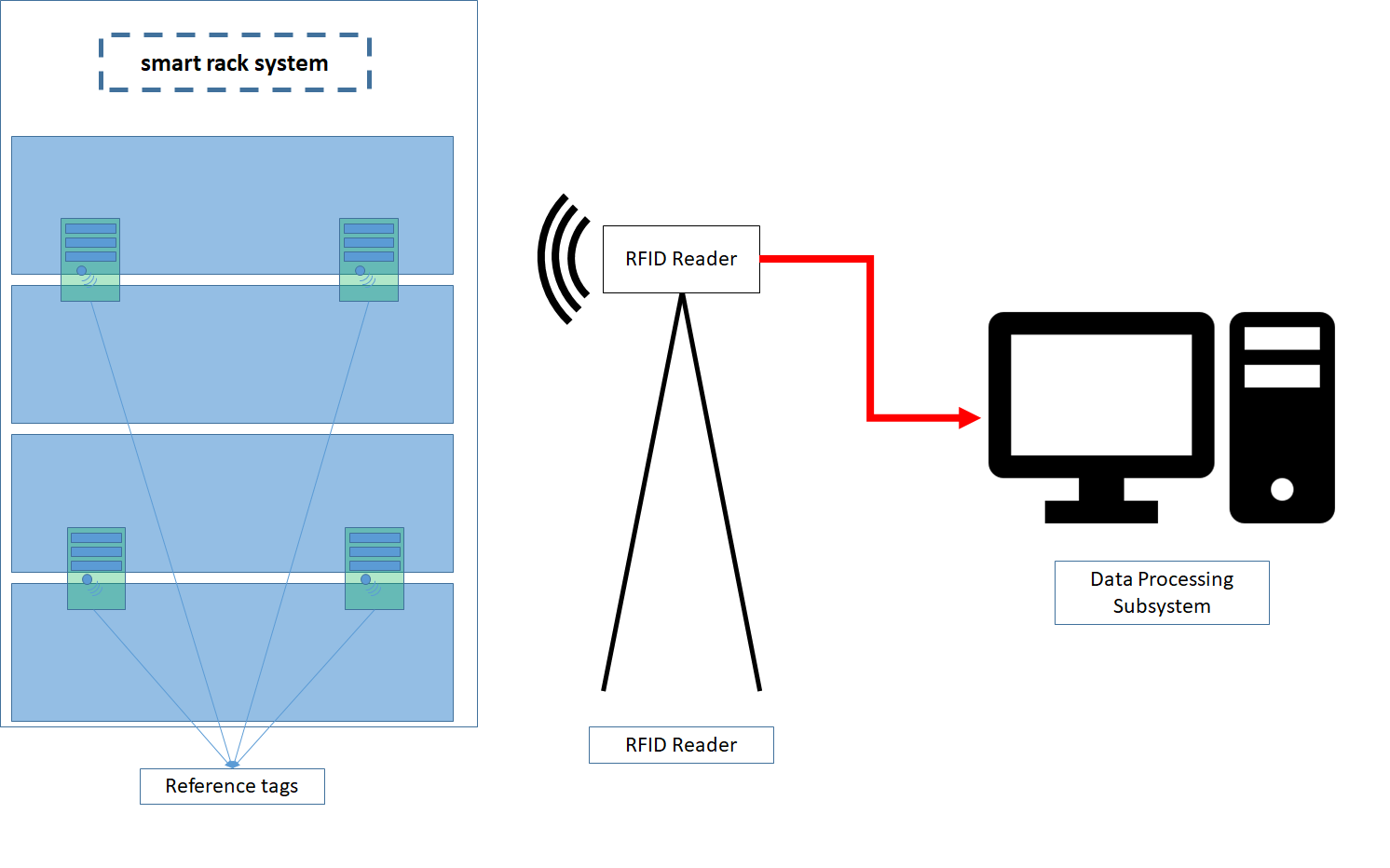}
			\par\end{centering}
		\caption{RFID localization testbed}
	\end{figure}
	\par\end{center} 

\subsection{RFID Dataset}

RFID-based localization has gained increasing popularity and is capable
of achieving milimetre accuracy. Furthermore, the use of COTS readers
feeding the phase information of the RF backscatter signal from the
RFID signal enhances object tracking accuracy. This approach offers
a cheaper solution than traditional approach in the object localization
involving deployment of antennas for every single object on a rack.This
problem presents the RFID-based localization in the manufacturing
shopfloor where the main goal is to track objects placed on the rack
or shells based on the phase information of the RFID tags (Courtesy
of Dr. Huang Sheng, Singapore). The application of machine learning
for the RFID based localization allows improvement of RFID-based localization
because the phase signal is heavily affected by the positional changes
of the objects. In addition, multiple items on the rack influences
the signal phase and amplitude distribution of a target tag. Online
learning is highly demanded in this application to solve the so-called
multipath effect due to reflections and scatterring distorting the
RSSI and phase of the signal. This problem is frequently encountered
positional configurations of the rack or shells or the sudden presence
of unknown objects. {Figure \ref{smartrack}} illustrates our test bed. 
\begin{table}[t]
		\caption{RFID Dataset}
		\begin{centering}
		\label{tab:RFID}
		\scalebox{0.8}{
		\begin{tabular}{lrrrrcl}
				\toprule
				\textbf{Model} & \textbf{Classification Rate} & \textbf{Node} &\textbf{Layer/Local Model} &\textbf{Runtime} &\textbf{Input} &\textbf{Architecture} \tabularnewline
				\midrule
				DSSCN & \textbf{0.99$\pm$0.02} &5.2$\pm$0.56 & 1.04$\pm$0.2 & 1.97$\pm$0.46 & 1 & Deep \tabularnewline
				eSCN &	0.92$\pm$0.08&18.69$\pm$11.92&\textbf{1}&	18.3$\pm$20.08&	1	&Shallow\tabularnewline
				Learn++NSE& \textbf{0.99$\pm$0.01}&	N/A&	100&	5.69$\pm$56.88&	1&	Ensemble\tabularnewline
				Learn++CDS&	\textbf{0.99$\pm$0.01}&N/A&	100	&7.99$\pm$9.92	&1&	Ensemble\tabularnewline
				pENsemble&	0.67$\pm$0.02&	2.02$\pm$0.2	&\textbf{1.01$\pm$0.07}	&\textbf{0.89$\pm$0.12}&	1&	Ensemble\tabularnewline
				pClass&	0.95$\pm$0.08&	\textbf{2}	&1	&1.13$\pm$0.25&	1&	Shallow\tabularnewline
				eT2Class&	0.94$\pm$0.04&	\textbf{2}	&1	&1.13$\pm$0.24&	1&	Shallow\tabularnewline
				\bottomrule 
		\end{tabular}}
		\par\end{centering}
\end{table}

The test bed consists of a steel storage rack with five compartments
where each compartment is occupied by 5-6 objects with different dimensions
attached with COTS RFID tags. Two UHF antennas are mounted on the
rack and connected to COTS RFID receiver. The RSSI and phase signals
reading of the COTS RFID receiver will be transmitted to the host
computer by means of Ethernet connections. The rack is divided into
four zones by the installment of four RFID reference tags in different
sections of the rack. The RFID dataset contains 281.3 K records of
RSSI and phase signals with four target classes corresponding to the
four zones of the rack. Our experiment is carried out using the periodic
hold out protocol with 100 time stamps. In each time stamp, 2 K data
samples are exploited for model updates, while the rest 813 data points
are fed to examine model's generalization potential. Consolidated
numerical results are summed up in Table 8. 

DSSCN produces promising generalization power attaining 99\% classification
rate. It is comparable to Learn++NSE and Learn++CDS but incurs less
complexity in terms of training speed and network complexities because
DSSCN realizes controlled growth of network structure where a new
data stream does not necessarily {trigger} addition of new building
units. The importance of the network depth is borne out in {Table 8}
where it delivers 7\% improvement on accuracies compared to its shallow
counterpart, eSCN. It is also shown that eSCN runtime is more sluggish
than those eT2Class and pClass because it has to fully utilize all
samples in the data chunk to execute the scope selection procedure. It is worth noting that the sliding window size in our numerical
study is fixed as the data chunk size. [-0.1,0.1], [-1,1], [-2,2], [-3,3] are selected by the SCN-based initialization strategy. Note that SCN-based initialization strategy is triggered on the fly whenever a new hidden node is appended. In other words, every hidden node are randomly crafted from dynamic random sampling scopes. Table 9 displays the ranges of random parameter produced by the SCN-based initialization strategy for all presented use cases. 
\begin{table}[t]
		\caption{Scopes of Random Parameters}
		\begin{centering}
		\label{tab:my_label}
		\scalebox{0.8}{
		\begin{tabular}{lc}
				\toprule
				\textbf{Problems} & \textbf{Ranges} \tabularnewline
				\midrule
				SEA & $[-0.5,0.5],[-1,1],[-2,2],[-3,3]$ \tabularnewline
                Weather & $[-100,100]$ \tabularnewline
                Electricity & $[-3,3]$ \tabularnewline
                SUSY & $[-100,100]$ \tabularnewline
                Hyperplane & $[-1,1],[-1.5,1.5],[-2.2],[-3,3]$ \tabularnewline
                RFID & $[-0.1,0.1],[-1,1],[-2,2],[-3,3]$ \tabularnewline
				\bottomrule 
		\end{tabular}}
		\par\end{centering}
\end{table}

\subsection{Prequential Test-then-Train Protocol}
This section discusses numerical validation of DSSCN in the prequential test-then-train scenario - another prominent evaluation procedure for data stream algorithms. Unlike the hold-out approach, the prequential test-then-train procedure first tests generalization performance of a model before being used for model updates. This scenario considers the fact that data stream often arrives without label and calls for operator effort to annotate the true class labels. There exists a delay factor which should be taken into account in getting access to the true class label because ground truth cannot be immediately obtained. The simulation protocol is exhibited in Fig. 2. As with earlier simulations, DSSCN is evaluated in five facets: classification rate, hidden nodes, hidden layer, runtime and input features and is simulated under this procedure for all the six problems. The number of time stamps are set as 240,91,200,566,36, respectively for hyperplane, electricity pricing, SEA, RFID, weather and SUSY problems. This setting leads each data batch to comprise 500 data points. This setting is meant to arrive at moderate-size data batch which properly examines algorithm's aptitude for concept drift. DSSCN is compared to a fixed deep neural network (DNN) to illustrate the benefit of stochastic depth to the learning performance. The number of hidden layer and node in the DNN are set such that it carries comparable complexity as that of DSSCN. In addition, DNN is run in a single-epoch thus being equivalent to an online learner with fixed capacity. Numerical results are reported in Table 10. 
\begin{table}[t]
		\caption{Prequential Test-Then-Train Procedure}
		\begin{centering}
		\label{tab:allresults}
		\scalebox{0.8}{
		\begin{tabular}{lrrrrc}
				\toprule
				\textbf{Model} & \textbf{Classification Rate} & \textbf{Node} &\textbf{Layer/Local Model} &\textbf{Runtime} &\textbf{Input} \tabularnewline
				\midrule
				SEA-DSSCN & 0.91$\pm$0.02 &\textbf{10.29$\pm$4.09} & 2.15$\pm$0.82 & 0.54$\pm$0.11 & 3 \tabularnewline
				SEA-DNN & \textbf{0.92$\pm$0.62} & \textbf{10} & \textbf{2} & \textbf{0.12$\pm$0.02} & 3\tabularnewline
				Hyperplane-DSSCN &	0.91$\pm$0.02&\textbf{12.96$\pm$5.18}&4.9$\pm$1.86&0.52$\pm$0.17&	4\tabularnewline
				Hyperplane-DNN & \textbf{0.92$\pm$0.26} & 19 & \textbf{4} & \textbf{0.18$\pm$0.02} & 4\tabularnewline
				Weather-DSSCN& \textbf{0.8$\pm$0.02}&	5.08$\pm$2.9&	2.4$\pm$1.27&	0.46$\pm$0.21&	8\tabularnewline
				Weather-DNN& 0.72$\pm$0.08 &\textbf{5} & \textbf{2} & \textbf{0.11$\pm$0.02} & 8\tabularnewline
				Electricity Pricing-DSSCN&	\textbf{0.68$\pm$0.13}&\textbf{3.58$\pm$1.45}&	3.58$\pm$1.45	&0.32$\pm$0.13	&8\tabularnewline
				Electricity Pricing-DNN& 0.57$\pm$0.11&\textbf{4}&\textbf{2}&\textbf{0.16$\pm$0.16}&8\tabularnewline
				RFID-DSSCN&	\textbf{0.99$\pm$0.03}&	\textbf{7.87$\pm$3.93}	&3.08$\pm$1.29	&0.33$\pm$0.11&	1\tabularnewline
				RFID-DNN&0.99$\pm$0.27&9&\textbf{2}&\textbf{0.11$\pm$0.01}&1\tabularnewline
				SUSY-DSSCN&	\textbf{0.77$\pm$0.02}&	\textbf{28.72$\pm$11.22}	&\textbf{10.62$\pm$3.88}	&\textbf{0.43$\pm$0.08}&	18\tabularnewline
				SUSY-DNN& 0.54$\pm$0.87 & 29 & 11 & \textbf{0.26$\pm$0.02} & 18\tabularnewline
				\bottomrule 
		\end{tabular}}
		\par\end{centering}
\end{table}

It is clear from Table 10 that learning performance of DSSCN is not compromised under the prequential test-then-train scenario. Moreover, the evolving nature of DSSCN is demonstrated where hidden layer and hidden nodes can be flexibly added and discarded. It is also noted that only the last layer is subject to model update during the stable phase. This setting is meant to expedite model updates without loss of generalization power since the last layer is supposed to craft the final predictive output. It is worth noting that DSSCN is structured under a deep stacked network architecture in which every layer takes the original input attributes mixed with a random shift.  

DSSCN also outperforms predictive accuracy of DNN in four of six problems featuring real-world dataset with substantial gap - almost 10$\%$ for weather problem and 20$\%$ for SUSY problem. This facet is possibly attributed by the interval type-2 fuzzy set of eSCN which handles noisy data better than the crisp activation function of neural network. Although DSSCN is inferior to DNN in other two problems, DSSCN still performs equally well where the difference of predictive accuracy between the two algorithms does not exceed $1\%$. Conclusive finding cannot be drawn in the context of hidden layer and hidden node because these two hyperparametets of DNN are selected such that they are on par to DSSCN. It is observed that DSSCN is slower than DNN in the execution time. DSSCN is built upon the stacked concept which operates in the classifier's level rather than in the layer's level, thereby calling for more expensive computational cost than that of DNN created under the multi-layer principle with a single optimization objective.
\subsection{{Ablation Study}}
This section outlines our numerical study which analyzes contribution of DSSCN's learning modules: hidden layer pruning, input weighting mechanism. It is done by activating and deactivating these learning modules which illustrates to certain extent both learning modules affect the learning performance. This numerical study makes use of SEA and hyperplane problems simulated under the prequential test-then-train procedure. The prequential test-then-train procedure is followed here because it offers more realistic data stream scenario than the periodic test-then-train mechanism because of the fact that data stream arrives with the absence of class labels. A classifier is supposed to generalize them first before receiving any updates. Numerical results are reported in Table 11. 

\begin{table}[t]
		\caption{Ablation Study}
		\begin{centering}
		\label{tab:Learning_Modules}
		\scalebox{0.8}{
		\begin{tabular}{lrrrrc}
				\toprule
				\textbf{Model} & \textbf{Classification Rate} & \textbf{Node} &\textbf{Layer/Local Model} &\textbf{Runtime} &\textbf{Input} \tabularnewline
				\midrule
				SEA-w/o FW & 0.88$\pm$0.11 & 10.29$\pm$4.09 & 11.16$\pm$4.61 &  2.34$\pm$0.88 & 3 \tabularnewline
				SEA-w/o LP & 0.91$\pm$0.05 & 14.38$\pm$6.03 & 3.19$\pm$1.3 & 0.45$\pm$0.1 & 3\tabularnewline
				Hyperplane-w/o FW &	0.88$\pm$0.1&10.74$\pm$4.23&2.34$\pm$0.91&0.54$\pm$0.12&	4\tabularnewline
				Hyperplane-w/o LP & 0.91$\pm$0.02 & 12.96$\pm$5.18 & 4.92$\pm$1.86 & 0.5$\pm$0.16 & 4\tabularnewline
				Weather-w/o FW& 0.69$\pm$0.2&	5.47$\pm$3.23&	2.42$\pm$1.32&	0.65$\pm$0.4&	8\tabularnewline
				Weather-w/o LP& 0.8$\pm$0.02&5.17$\pm$3.06 & 2.41$\pm$1.31 & 0.52$\pm$0.17 & 8\tabularnewline
				Electricity Pricing-w/o FW&	0.59$\pm$0.1&45.4$\pm$23.09&	3.58$\pm$1.45	&1.89$\pm$2.48	&8\tabularnewline
				Electricity Pricing-w/o LP& 0.68$\pm$0.13&3.58$\pm$1.45&3.58$\pm$1.45&0.39$\pm$0.13&8\tabularnewline
				RFID-w/o FW&	0.91$\pm$0.15&	10.18$\pm$5.2	&3.08$\pm$1.27	&0.55$\pm$0.22&	1\tabularnewline
				RFID-w/o LP&0.99$\pm$0.02&7.87$\pm$3.93&3.08$\pm$1.27&0.6$\pm$0.25&1\tabularnewline
				SUSY-w/o FW&	0.57$\pm$0.08&	37.25$\pm$12.78	&20.2$\pm$11.12	&0.25$\pm$0.13&	18\tabularnewline
				SUSY-w/o LP&	0.75$\pm$0.08&	47.97$\pm$26.03	&20.2$\pm$11.12	&0.37$\pm$0.11&	18\tabularnewline
				\bottomrule 
		\end{tabular}}
		\par\end{centering}
\end{table}

From Table \ref{tab:Learning_Modules}, the effect of online feature weighting and hidden layer pruning mechanims are clearly demonstrated. The absence of online feature weighting results in substantial drop in predictive accuracy in all six cases. Although the evolution of feature weights reveals that feature weights revolve around 1, they remain important to DSSCN to produce decent numerical results. Note that the feature weights functions not only to exclude unimportant input features but also to improve feature representation via the weighting mechanism. The network compression via the layer pruning component is proven to be effective for complexity reduction mechanism {while still retaining high predictive accuracy.} It is seen that the network complexity increases when deactivating this learning component.

\section{Reason Behind Deep Structure for Data Streams}
This section elaborates the rationale behind the requirement of deep approach in data stream context and why addition of hidden layer is more effective than addition of hidden nodes in the presence of concept drift to resolve the underfitting situation due to drift. The rationales are outlined in the sequel.
\begin{itemize}
    \item \textit{Power of Depth}: the advantage of depth in the feed-forward network has been proven not only numerically but also theoretically in the literature. \cite{powerofdepth}, 
    It points out that addition of even only one hidden layer can be exponentially more valuable than width (neuron, rule, etc) for standard feed-forward neural networks.
    
An in-depth analysis of the contribution of depth to learning performance is also studied in \cite{deeplinear} in terms of the number of linear regions generated by deep neural network (DNN). In p.2 of \cite{deeplinear}, it is mentioned that {the capacity to replicate computations over the input space grows exponentially with the number of network layers}. In Theorem 5 of \cite{deeplinear}, it is seen that the number of linear regions computed by a neural network is a factor of the number of hidden layers. This finding reports the capacity of neural networks. 

This result is further confirmed in \cite{linearregion2}. It is said in the abstract that {deep neural network has considerably much more linear regions than that of a shallow network}. This implies the higher the number of hidden layers the better the capability in producing more complex functions because it can use information of previous layers. Nevertheless, this is subject to the number of available samples (p. 2 of \cite{linearregion2}) because it increases the risk of overfitting. As known from the popular bias-variance dilemma, overfitting can be reduced by feeding {additional samples to the network.} 
\item \textit{Bias-and-Variance Dilemma}: in the presence of drift, it can be simply said that a model suffers from an underfitting issue because it deals with unexplored regions. This issue incurs high predictive error due to the high bias problem. The appropriate step in handling the high bias problem (underfitting) is by increasing network complexity. It can be done by adding rules, neurons, classifiers as well as hidden layers. This is the main reason why addition of neurons, rules or classifiers works well in handling concept drifts. In our recent work \cite{DEVDAN}, we address non-stationary environments by devising the bias-variance formulas as hidden unit growing and pruning criteria. As the finding of \cite{powerofdepth}, addition of new layer is more effective not only in resolving the high bias problem as a result of concept drift but also in improving the generalization power in the long term. This fact is recently confirmed in \cite{verydeep} where increasing the network depth has shown significant performance improvement although a small filter size is used in the CNN context. 
\item \textit{Stochastic Depth}: it is well-known that the main bottleneck of deep neural network (DNN) is found in its static structure and {in the difficulty} in estimating the proper network complexity without performing some offline mechanism. This forces one to simply go to a very deep network structure which usually gives decent approximation power \cite{verydeep}. However, a very deep network architecture also faces its own set of challenges: exploding and vanishing gradient, diminishing feature reuse in addition to expensive computational and memory complexities. The concept of stochastic depth itself as adopted in our work is not new. In \cite{stochasticdepth}, the dynamic depth network is proposed where the key idea is to start with a very deep network architecture and drops a subset of hidden layers and bypass them with the identity function. The key difference with our approach lies in the construction of network structure where we starts from a shallow network with only a single generalizer (classifier) while gradually deepening the network structure if concept drift is detected. 

The key is only to assure that addition of network complexity does not induce the catastrophic forgetting problem where previously valid knowledge is erased by a new one. In realm of multi-layer network (not stacked), addition of hidden layer when a drift occurs causes the catastrophic forgetting problem because the final prediction heavily relies on the mechanics of last layer. This problem does not apply in the context of stacked network structure, because each layer is formed by a local learner (classifier) and has its own generalization power. Every layer minimizes its own cost function. 
\item \textit{Concept Drift Detection}: the data stream problem has started to attract research attention in the deep learning area, usually framed under the roof of continual learning (https://www.continualai.org/). Most approaches deal with data streams in the context of multi-task learning where incoming data batch is of different tasks. In \cite{ECW}, the idea of elastic weight consolidation is introduced. This approach adds new neurons for each incoming task and EWC method is designed to assure that weigh of new tasks should not be deviated too far of the old ones to address the issue of catastrophic forgetting. In \cite{PNN}, addition of some nodes for every layer is carried out for each task and the catastrophic forgetting is handled by freezing the old parameters. These two approaches impose prohibitive space complexity because the network capacity grows uncontrollably. \cite{dynamicallyexpandable} offers advancement of the two approaches via the selective retraining and growing approach. These approaches select relevant parts of old neural networks to be applied in the new tasks. Moreover, $k$ hidden nodes are added based on the loss function criteria. 
The drift detection method enables us to estimate the novelty of data stream in respect to the old ones. Significant difference of new data stream triggers insertion of new hidden layer which increases capacity of network structure to accommodate new concepts. Note that concept drift is evaluated for each data batch rather than data points. This strategy aims to have sufficient evaluation window to declare the concept drift situation thereby preventing the false alarm. Moreover, DSSCN is equipped by the rule growing scenario to overcome local change (per data point). On the other hand, the catastrophic forgetting problem is dealt with the freezing mechanism, only the last layer is tuned. In addition, DSSCN is built upon a stacked network structure which consists of a collection of classifiers connected in tandem minimizing their own cost functions. To sum up, the use of drift detection mechanism as a way of hidden layer growing mechanism is to set every hidden layer as representation of different concepts in the data streams.
\item \textit{Numerical Results}: additional numerical results of static DNNs have been added in section as a benchmark to show performance improvement of our approach. Simulation is carried out under the prequential test-then-train procedure. It is reported that DSSCN outperforms static DNN in terms of predictive accuracy.  
\end{itemize}

\section{Conclusions}
The advantages of DSSCN has been validated using six real-world
and synthetic datasets featuring non-stationary characteristic. Furthermore,
DSSCN has been numerically validated under two standard evaluation protocols of data stream learner, periodic hold-out and prequential test-then-train where it performs equally well under the two scenarios and demonstrates
state-of-the art performance compared to five data stream algorithms and fixed-size DNN. The contribution of learning components has been numerically analyzed in which each of them supports the learning performance of DSSCN.  
It is worth noting that eSCN has no tuning mechanism of hidden nodes
rather random parameters are sampled from adaptively selected scope.
The random parameter constraint imposes the use of all data points in the data chunk during the creation of new hidden unit which slightly slows down model updates but remains competitive to existing data stream methods.
This result is mainly caused by the update strategy of DSSCN where only the last layer is fine-tuned during the stable condition to generate different input representations and to overcome the catastrophic forgetting problem. In addition, this update strategy is confirmed by the relevance of the last layer because it is supposed to represent the closest concept to the original data distribution. That is, it is added when the latest concept change is signalled.  

Although the drift detection mechanism has been demonstrated to be
very effective in growing the depth of deep neural networks, it does
not evaluate model's generalization. It is well-known that the depth
of deep neural networks has direct connection to the generalization power.
Our future work will be focused on the algorithmic development of
generalization-based criteria to evolve the structure of deep neural
networks. Also, we will investigate the application of proposed
methodologies to other deep learning variants including CNN and LSTM in our future studies. 

\section{Acknowledgement}
This work is fully supported by Ministry of Education, Republic of Singapore, Tier 1 Research Grant and NTU Start-up Grant. The authors also thank Singapore Institute of Manufacturing Technology (SIMTech), Singapore for providing RFID dataset and acknowledge the assistance of Mr. MD. Meftahul Ferdaus for Latex typesetting of this paper. 

\bibliographystyle{unsrt}  
\bibliography{references}

\end{document}